\documentclass[10pt, a4paper, twocolumn, showabstract]{naverlabseurope}

\usepackage{times}
\usepackage{latexsym}
\usepackage[T1]{fontenc}
\usepackage[utf8]{inputenc}
\usepackage{microtype}
\usepackage{inconsolata}
\usepackage{graphicx}
\usepackage{subcaption}
\usepackage{tabularx}
\usepackage{booktabs}
\usepackage[dvipsnames,table]{xcolor}
\usepackage{colortbl}
\usepackage{color}
\usepackage{amsmath}
\usepackage{amssymb}
\usepackage{hyperref}
\usepackage{adjustbox}
\usepackage{pifont}
\usepackage{pgf}
\usepackage{multirow}
\usepackage{enumitem}

\title{Ready to Speak: Aligning LLMs for TTS-Friendly Text Generation}
\titlerunning{Ready to Speak: Aligning LLMs for TTS-Friendly Text Generation}

\correspondingauthor{thibaut.thonet@naverlabs.com}

\authors{Thibaut Thonet, Jos Rozen, Laurent Besacier}
\affiliations{NAVER LABS Europe}
\contributions{}
\website{\texttt{\{thibaut.thonet,jos.rozen,laurent.besacier\}@naverlabs.com}}
\websiteref{}

\newcommand{\colorvaluegradient}[1]{%
  \pgfmathsetmacro{\val}{int(min(abs(#1)*110,100))}%
  \pgfmathparse{#1 > 0 ? "red!\val!white" : (#1 < 0 ? "blue!\val!white" : "white")}%
  \expandafter\cellcolor\expandafter{\pgfmathresult}{#1}%
}

\begin{abstract}
Current Large Language Models (LLMs) are primarily optimized for written text, often producing outputs that are grammatically correct and helpful yet poorly suited for spoken delivery via Text-to-Speech (TTS). In this work, we study how to make LLMs \textit{natively} generate TTS-friendly text, which we frame as a preference alignment problem: instead of relying on downstream rewriting modules, we directly align LLMs to generate text optimized for spoken delivery. We introduce two preference datasets spanning different target domains, CORA and Recipe, which contain paired TTS-friendly and TTS-unfriendly responses. We further propose an evaluation suite combining a pattern-based heuristic metric, a TTS$\to$ASR evaluation pipeline, and a MUSHRA listening study with human judges. Our experiments compare the recently proposed Feature-aware Sampling and Tuning (FaST) framework~-- leveraging interpretable features instead of a black-box reward model~-- against an array of alignment baselines on the TTS-friendly generation task. Notably, we found that FaST achieves the best overall tradeoff between TTS-friendliness and helpfulness across various settings. We also identified a strong correlation between our different metrics, highlighting the ability to reliably assess TTS-friendliness via an efficient heuristic.
\end{abstract}

\begin{document}
\maketitle

\section{Introduction}
\label{sec:introduction}

Currently, LLMs are primarily trained using textual data and preferences, which results 
in grammatically correct text that however might not be optimized for spoken delivery 
by Text-to-Speech (TTS) systems. The generation of text suitable for TTS synthesis 
presents unique challenges beyond traditional natural language generation tasks. Text must be optimized not only for semantic correctness and fluency, but also for 
characteristics that govern spoken delivery~-- which we refer to as TTS-friendliness.

We frame TTS-friendly text generation as a \textit{preference alignment} problem: 
rather than relying on post-hoc rewriting modules to perform so-called \textit{text normalization}~\citep{Zhang2024tn,Wong2025tn}, we directly align an LLM to produce 
outputs ready for spoken delivery. This is practically motivated: post-processing 
adds noticeable latency, requires complete sentences before it can run, and is 
inherently tied to the specific tokenization and input format of each TTS system. 
Producing TTS-friendly outputs directly simplifies the architecture, reduces latency, 
lowers inference costs, and yields text that can be consumed by any TTS system 
out of the box.

Concretely, we define TTS-friendliness through a short, interpretable rule set and 
measurable objectives, and study how well existing alignment techniques can satisfy it. A key challenge is that TTS-friendliness is \textit{multi-objective}: a 
response must avoid TTS-unfriendly surface forms (symbols, abbreviations, raw URLs, 
compact numeric notation) while remaining factually correct and helpful. We show that 
preference alignment methods can effectively address this challenge, even in low-data 
regimes with as few as 10 training examples.

The contributions of this paper are as follows. \textbf{(1) Problem framing:} we 
formalize TTS-friendly text generation as a preference alignment problem, with a 
concise rule set and measurable objectives. \textbf{(2) Dataset curation:} we 
introduce two preference datasets: CORA, a synthetic \underline{C}offee 
\underline{OR}dering \underline{A}ssistant benchmark targeting conversational 
responses, and Recipe, derived from a large cooking recipe corpus targeting 
procedural descriptions~-- each containing paired TTS-friendly and TTS-unfriendly 
responses covering symbols, abbreviations, numeric shorthands, and raw URLs. 
\textbf{(3) Metric suite:} we propose and validate a heuristic score and a 
TTS$\to$ASR round-trip pipeline, confirming their agreement with human judgments 
via a MUSHRA listening study. \textbf{(4) Method comparison:} we adopt 
FaST~\cite{fast}, replacing black-box reward models with interpretable features, 
and compare it against prompting, SFT, DPO, GRPO, and RFT on two datasets and 
two data regimes (10 vs.\ 100 samples). FaST achieves the best 
TTS-friendliness/helpfulness tradeoff, outperforming baselines with 
as few as 10 examples.\footnote{Our datasets, metrics and code are publicly available at \url{https://github.com/naver/tts-friendly-gen}.}

The rest of the paper covers related work (Section~\ref{sec:related}), datasets and 
metrics (Section~\ref{sec:data}), method and baselines (Section~\ref{sec:method}), 
experimental setup and results (Sections~\ref{sec:experiments}--\ref{sec:results}), 
metric validation (Section~\ref{sec:validation}), and conclusion (Section~\ref{sec:conclusion}).
\section{Related Work}
\label{sec:related}

\subsection{LLM Alignment and Control for Speech Generation}

A foundational approach to generating speech-friendly text is instruction-tuning LLMs
specifically for spoken delivery. \citet{speechworthy2024} introduce the concept of
``speechworthy'' text generation, demonstrating that LLMs can be tuned to produce
text optimized for TTS applications. Separately, preference alignment has been applied
to improve the \textit{acoustic} outputs of TTS systems themselves: \citet{tian2025preference}
show its benefits for LLM-based TTS decoders; \citet{hussain2025koel} combine it with
classifier-free guidance; and \citet{zhang2025advancing} leverage it to enhance
zero-shot TTS intelligibility across challenging domains. However, these works
optimize the TTS component and rely on human auditory preference datasets, leaving
the \textit{text} side of the pipeline largely unaddressed. In contrast, we focus on
aligning the LLM to natively generate TTS-friendly text, independently of the
downstream TTS system.

Beyond alignment, \citet{samuel2025cie} propose continuous signals to control LLM
generation for fine-grained spoken characteristics, sharing our focus on the text
generation side though requiring architectural modifications. Domain-specific
challenges are addressed by \citet{mathreader2024} for mathematical documents, while
\citet{xie2024controllable} provide a broader survey of controllable speech synthesis
in the LLM era.

\subsection{Constrained Text Generation and Captioning}

Constrained text generation is relevant to TTS-friendly generation, as both require
satisfying explicit requirements on form while preserving content.
\citet{xie2025shotbyshot} generate audio descriptions under strict constraints using
a training-free VLM-based approach. \citet{hirsch2021clid} address length control in
image captioning with limited data via constrained decoding strategies.
\citet{cardei2025constrained} integrate hard constraints directly into the denoising
process of discrete diffusion models, providing inspiration for constraint-aware text
generation. More broadly, \citet{duran2025beyond} show that optimizing text for a
specific consumption context improves utility beyond fluency alone.

\subsection{Discussion}

The works reviewed above highlight two gaps. First, existing alignment work
predominantly targets the acoustic output of TTS systems, treating the upstream text
as fixed. \citet{speechworthy2024} is a notable exception, yet their work addresses
broad speech-suitability rather than the fine-grained properties (numeric formatting,
abbreviation density, conversational tone) that determine reliable synthesis across
TTS engines. Moreover, their human annotations conflate helpfulness and
speechworthiness, making it difficult to isolate the contribution of each dimension.
We address both limitations by framing TTS-friendly text generation as a preference
alignment problem with explicit, disentangled objectives, operating upstream of any
TTS system. Second, existing approaches rely on costly and subjective human auditory
judgments. Constraint-aware methods~\citep{cardei2025constrained, hirsch2021clid}
show that interpretable objectives can guide generation effectively, but have not
been applied to TTS-friendliness. We bridge this gap by grounding alignment in
interpretable text-level features that serve simultaneously as evaluation metrics
and as a guide to build synthetic preference datasets, enabling multi-objective
control in a low-data regime.
\section{Datasets \& Evaluation Metrics}
\label{sec:data}

\paragraph{Datasets}

We conduct our study on two datasets spanning distinct domains and text styles.
\textbf{CORA} is a synthetic question-answering dataset for a coffee-shop assistant,
targeting conversational responses and constructed to highlight several types of
TTS-unfriendly content (symbols, acronyms, prices, time expressions, email addresses,
URLs, etc.). \textbf{Recipe} is based on the publicly available RecipeNLG\footnote{\url{https://huggingface.co/datasets/mbien/recipe_nlg} (restricted to non-commercial usage only)} dataset~\cite{bien2020recipenlg}
(originally containing 2.2M cooking recipes), targeting procedural descriptions.
From this corpus, we sample a pool of 300 recipes. Each instance in both datasets
includes a TTS-friendly (\emph{chosen}) and a TTS-unfriendly (\emph{rejected})
response, used for preference alignment. For every dataset, we generate five random
train/validation/test splits from this shared pool; the per-fold sizes are listed
in Table~\ref{tab:datasets}.
Sizes are deliberate: they match a deployment scenario with only a few
preference tuples per domain and, as no public benchmark existed, both
datasets had to be built from scratch. Synthetic-corpus diversity also
saturates beyond a few hundred items, so larger test pools would add
cost without sharpening the evaluation. App.~\ref{app:examples}
(Table~\ref{tab:appendix_examples}) shows one illustrative
$\langle$context, chosen, rejected$\rangle$ tuple per dataset.

\begin{table}[t]
  \centering
  \small
  \begin{tabular}{lrrrr}
    \toprule
    \textbf{Dataset} & \textbf{Total} & \textbf{Train} & \textbf{Validation} & \textbf{Test} \\
    \midrule
    CORA   & 262 & 100 & 81  & 81  \\
    Recipe & 300 & 100 & 100 & 100 \\
    \bottomrule
  \end{tabular}
  \caption{Number of distinct preference tuples in each dataset and per-fold sizes. The same pool is reshuffled across 5 folds (Train + Val + Test = Total in each fold).}
  \label{tab:datasets}
\end{table}


\paragraph{Metrics}

Our experimental evaluation jointly quantifies task helpfulness and TTS-friendliness, and we analyze the trade-off between these two objectives throughout the paper. We deliberately measure helpfulness on the target tasks rather than general-purpose capabilities, which matter less for the domain-specific use cases we consider (e.g., coffee-shop assistance, cooking guidance).

Our \emph{heuristic TTS-friendliness} metric is a pattern-based score on a 1--5 scale (higher meaning
more TTS-friendly) that penalizes surface-level TTS-unfriendly elements such
as symbols, abbreviations, URLs, email addresses, and poorly formatted prices,
times, or quantities. The exact regular expressions, weights, length
normalization, and 1--5 mapping are detailed in
App.~\ref{app:heuristic-metric}.

The \emph{TTS$\to$ASR} metric measures TTS-friendliness through a synthesis-transcription round-trip: a TTS-friendly text should be trivially speakable and its audio should transcribe back to the original text. Concretely, each response is synthesized with Kyutai Pocket TTS~\cite{kyutai_pockettts} (voice \textit{alba}, 24~kHz, on CPU) and transcribed back with Wav2Vec2-large~\cite{wav2vec2_large_960h_lv60_self}. Character and Word Error Rates (CER, WER) are then computed against the original text; lower scores indicate more TTS-friendly text, as problematic elements are expected to degrade synthesis and increase transcription errors.

A \emph{Helpfulness} metric on a 1–5 scale, assessed by GPT-4o-mini, complements the TTS-friendliness score by evaluating factual correctness and the extent to which the user's request is addressed, independently of stylistic considerations.
The metric follows the rubric-based LLM-judge paradigm \cite{kim-etal-2024-prometheus-1,kim-etal-2024-prometheus,thonet-etal-2025-elitr}, which has been shown to correlate strongly with human judgments when models are provided with explicit scoring criteria. For the Recipe dataset, we use the preferred response from each preference tuple (i.e., the TTS-friendly variant) as a reference in the judge prompt, thereby anchoring the 1–5 helpfulness score. In contrast, CORA questions are more open-ended, making it difficult to define a single canonical reference answer; we therefore adopt a reference-free evaluation. To ensure that judgments remain well-grounded in this setting, the prompt first decomposes the user request into a set of atomic requirements, evaluates the degree to which each requirement is satisfied by the candidate response, and only then assigns an overall 1–5 helpfulness score. Full prompt templates are provided in App.~\ref{app:prompts}.

\section{Method}
\label{sec:method} 

We frame TTS-friendly text generation as a preference alignment problem in which an LLM must generate responses that are optimized for spoken delivery by downstream TTS systems. We also wish that the generated answers preserve their helpfulness. Rather than relying on common alignment frameworks based on traditional, monolithic reward models, we adopt the recently proposed Feature-aware Sampling and Tuning framework \citep{fast}~-- FaST~-- to tackle TTS-friendly generation. FaST was originally introduced for personalized preference alignment in a low-data regime, using interpretable high-level features and a lightweight reward formulation. 

Our adoption of FaST is motivated by three observations. First, many properties that influence TTS-friendliness~-- such as abbreviation usage, numeric formatting, sentence structure, or conversational tone~-- can be described through explicit and interpretable features. Second, TTS-friendliness is inherently multi-objective: improving listenability should not excessively degrade informativeness or helpfulness. Third, it may be costly and challenging to obtain a large-scale preference dataset that reflects TTS-friendly and TTS-unfriendly behaviors in the domain of interest.
FaST naturally supports this setting by leveraging a parameter-scarce Feature-aware Reward Model defined as a weighted combination of automatically discovered, interpretable features~-- rather than relying on a traditional, black-box reward model that is often overparameterized for low-data settings.

\subsection{Feature-aware Sampling and Tuning}
\label{sec:fast}

Our method follows the FaST framework~\citep{fast} which operates in four steps: (i) feature discovery, (ii) feature-wise response scoring, (iii) feature weight learning, and (iv) sampling-and-tuning alignment.

\paragraph{Feature discovery.} 
First, a set of high-level features is automatically discovered from the questions and associated options in the preference dataset. In practice, this is done by simply prompting an LLM with all the preference tuples (or a subset of them) in its context and requesting a list of $F$ features that globally capture the contrasting properties of the different options associated to the same question. 
This yields high-level descriptors that capture salient stylistic and structural properties of the text. This procedure also has the benefit of providing highly interpretable features that facilitate transparency. We show in Table~\ref{tab:cora-features10} a sample of the features discovered by applying this methodology to CORA; the full set of features for CORA and Recipe is also detailed in App.~\ref{app:features}.

\paragraph{Feature-wise response scoring.} 
Once the features are identified, each response in the training preference set is scored independently along every feature dimension using LLM-based feature functions $\{\phi_f\}_{f=1}^F$. For a question $x$ and a response $y$, $\phi_f{(x, y)}$ returns a score between 1 and 5 describing how prevalent feature $f$ is for response $y$ in the context of question $x$. In practice, $\phi_f$ is implemented by prompting an LLM using the question, the response and the description of the feature generated in the discovery step. Importantly, $\phi_f$ is frozen and thus require no parameter learning.

\paragraph{Feature weight learning.} 
Given the feature-wise scores, the Feature-aware Reward Model (FaRM) is defined as a linear combination of feature functions: $R_{\text{FaRM}}(x, y) = \sum_{f=1}^F  \lambda_{f} \, \phi_{f}(x, y)$ where $\lambda_f$ is the learned weight associated with feature $f$. Unlike traditional reward models, FaRM contains only a very small number of learned parameters~-- one scalar per feature. This makes the method particularly attractive in low-data settings and provides strong interpretability. The feature weights $\{\lambda_f\}_{f=1}^F$ are learned from preference annotations, maximizing the likelihood of assigning higher rewards to preferred responses using a Bradley-Terry style objective. This formulation allows FaRM to explicitly capture trade-offs between the competing objectives implicitly expressed in the preferences. For instance, in the context of CORA, the learned reward simultaneously favors
conversational phrasing and spelled-out numbers while penalizing excessive
abbreviations or symbol usage.

\paragraph{Alignment via sampling and tuning.}
The final step of FaST consists in fine-tuning an LLM so that it generates high-reward outputs. The approach used consists of an iterative sampling-and-tuning procedure that repeats three steps: (i) the model samples multiple candidate responses for each prompt; (ii) candidate responses are ranked using FaRM; (iii) the generation model is fine-tuned using the highest-ranked responses, via either DPO~\citep{dpo} or supervised fine-tuning (SFT) . The former yields Online-DPO, and the latter Rejection-sampling Fine-Tuning (RFT)~-- our focus here, as it was shown to give strong results in~\citet{fast}.

\subsection{Baselines}

We compare FaST against a range of baselines covering prompting, supervised fine-tuning, reward-model-free preference alignment as well as alignment based on a traditional reward model.

\begin{itemize}
    \item \textbf{Zeroshot}: The base LLM is directly used with a prompt that only describes the task (coffee-shop assistance and cooking recipe description for CORA and Recipe, respectively) without mentioning the TTS-friendliness requirement.
    \item \textbf{Prompting}: The base LLM is prompted with the task description and a list of requirements seeking to boost the outputs' TTS-friendliness. The prompt used is detailed in App.~\ref{app:prompts}.
    \item \textbf{SFT}: The model is fine-tuned on the preferred (``chosen'') responses from the preference datasets using standard supervised learning.
    \item \textbf{DPO}: The model is aligned with Direct Preference Optimization~\citep{dpo} using pairwise preferences between chosen and rejected responses, without training an explicit reward model.
    \item \textbf{GRPO-RM}: The model is fine-tuned via the Group Relative Policy Optimization technique~\citep{grpo} using a traditional reward model trained beforehand on the preference data.
    \item \textbf{RFT-RM}: This approach is a rejection-sampling fine-tuning baseline following the same sampling-and-tuning procedure as FaST, but using a traditional reward model instead of the feature-based FaRM. The reward model is the same as for GRPO-RM.
\end{itemize}

Additionally, we define the \textbf{Oracle} as the approach that returns the TTS-friendly (i.e., ``chosen'') responses directly from the preference datasets. This serves as an approximate upper bound on achievable performance.
\section{Experimental Setup}
\label{sec:experiments}

Our experiments study the performance of the compared approaches with respect to 
the proposed heuristic TTS-friendliness metric\footnote{The heuristic metric is 
validated as correlating well with the more complex TTS$\to$ASR metric as well as 
human judgments, as detailed in Section~\ref{sec:validation}.} as well as the 
Helpfulness metric based on an LLM-judge. We conduct our experiments on the CORA and Recipe datasets introduced in this paper and consider two data regimes: using only 10 training samples or the full training set (100 samples). These two settings were defined to roughly match the low-data scenarios introduced in \citet{fast}. The validation and test sets used for the evaluation both contain 81 samples for CORA and 100 samples for Recipe. Each experiment is replicated 5 times on a different train/validation/test split, and one response is generated per context. Every reported number averages over validation and test sets across all 5 folds~-- effectively 162 (CORA) and 200 (Recipe) utterances per fold~-- resulting in a total of 810 and 1000 utterances evaluated per approach, respectively. 

We use Qwen3-4B\footnote{\url{https://huggingface.co/Qwen/Qwen3-4B-Instruct-2507}} as the base model in our experiments (and also report results for SmolLM3-3B in App.~\ref{app:results-smollm}), with its reasoning abilities disabled. Qwen3-4B was adopted as well to implement the feature functions of FaST. In this approach, we used 40 features,\footnote{The number of features was set based on the two settings discussed in~\citet{fast} ($F=20$ and $F=40$) and the observation in pilot experiments that $F=40$ produced slightly richer and more informative feature descriptions. Importantly, this choice was made before running any generation experiments, so as to avoid any unfair advantage given to FaST over competing approaches.} that were discovered by prompting GPT-5.1\footnote{\url{https://developers.openai.com/api/docs/models/gpt-5.1}} with the original feature discovery prompt proposed by~\citet{fast}. The context of the prompt integrated the training set~-- either 10 or 100 samples depending on the setting. The hyperparameters of the different approaches are described in App.~\ref{app:hyperparams}.

\section{Results}
\label{sec:results}

\subsection{Preference Prediction Results}
\label{sec:response-prediction}

We first compare the ability of the feature-aware reward model (FaRM) and that of the traditional reward model (RM) to predict the preferences (i.e., the chosen response over the rejected one) on CORA and Recipe. The results on the full and 10-sample training sets are reported in Table~\ref{tab:preference-prediction}. The numbers correspond to the average of the validation and test accuracies, over the 5 splits. The overall high accuracy suggests that the TTS-friendly response is relatively easy to distinguish from the TTS-unfriendly one. This confirms that the preference datasets provide a strong training signal. In most cases, FaRM and RM obtain comparable accuracies~-- with however only 40 learned parameters for the former and 4B for the latter. We nonetheless note that FaRM was able to maintain a 0.9+ accuracy in the 10-sample setting of CORA while RM got substantially degraded compared to the full training setting. This is likely due to FaRM's high-parameter efficiency compared to RM.

\begin{table}[t]
    \centering
    \scalebox{0.9}{
    \begin{tabular}{l lcc}
    \toprule
    \multirow{2}{*}{\textbf{Approach}} & \multirow{2}{*}{\textbf{Train size}} & \textbf{CORA} & \textbf{Recipe} \\
    & & \textbf{Acc. $\uparrow$} & \textbf{Acc. $\uparrow$} \\
    \midrule
    RM   & 100 (full) & 0.996 & 0.998 \\
    FaRM & 100 (full) & 0.984 & 0.914 \\
    \midrule
    RM   & 10 & 0.871 & 0.946 \\
    FaRM & 10 & 0.960 & 0.941 \\
    \bottomrule
    \end{tabular}
    }
    \caption{Preference prediction results for reward models trained on the full training set or from 10 training samples. The reported numbers correspond to the average of the validation and test accuracies (higher is better).}
    \label{tab:preference-prediction}
\end{table}

\subsection{Generation Results}

\begin{figure*}[t]
    \centering
    \begin{subfigure}[b]{0.49\textwidth}
        \includegraphics[width=\textwidth]{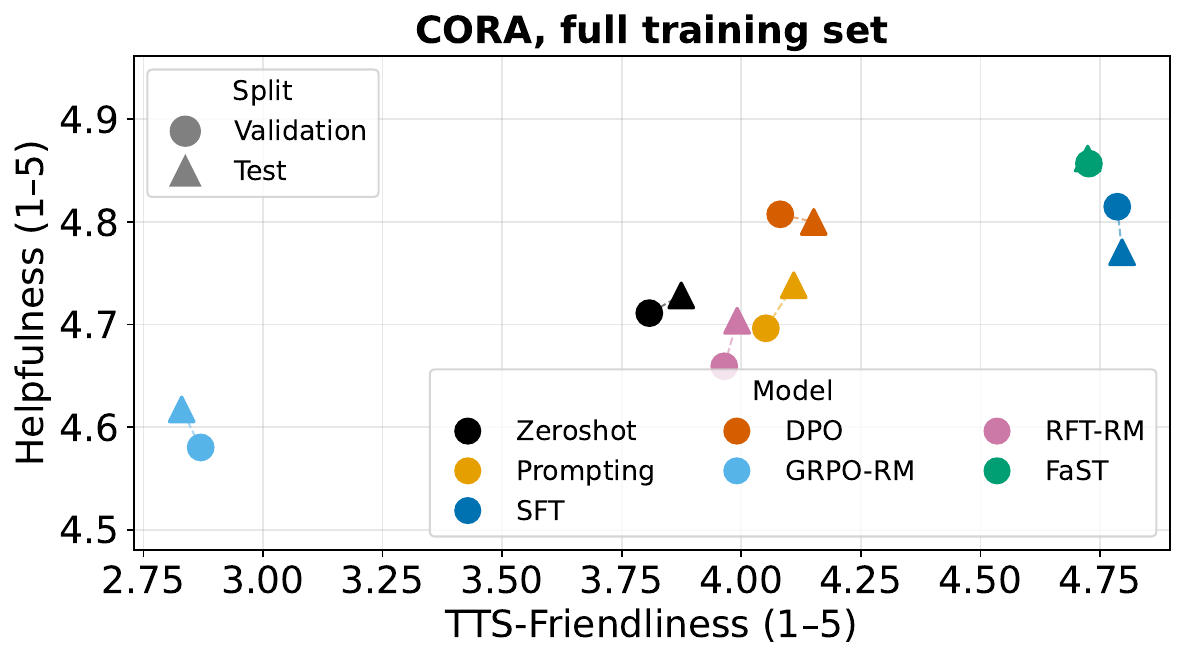}
    \end{subfigure}
    \hfill
    \begin{subfigure}[b]{0.49\textwidth}
        \includegraphics[width=\textwidth]{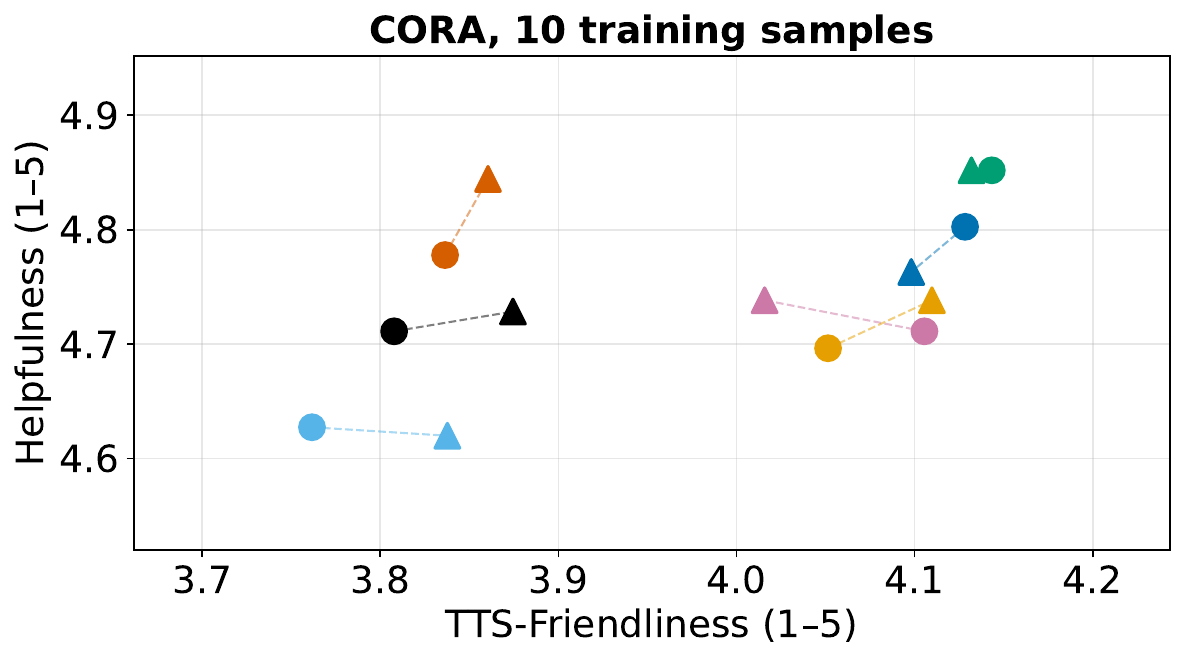}
    \end{subfigure}
    \vspace{0.1cm}
    \begin{subfigure}[b]{0.49\textwidth}
        \includegraphics[width=\textwidth]{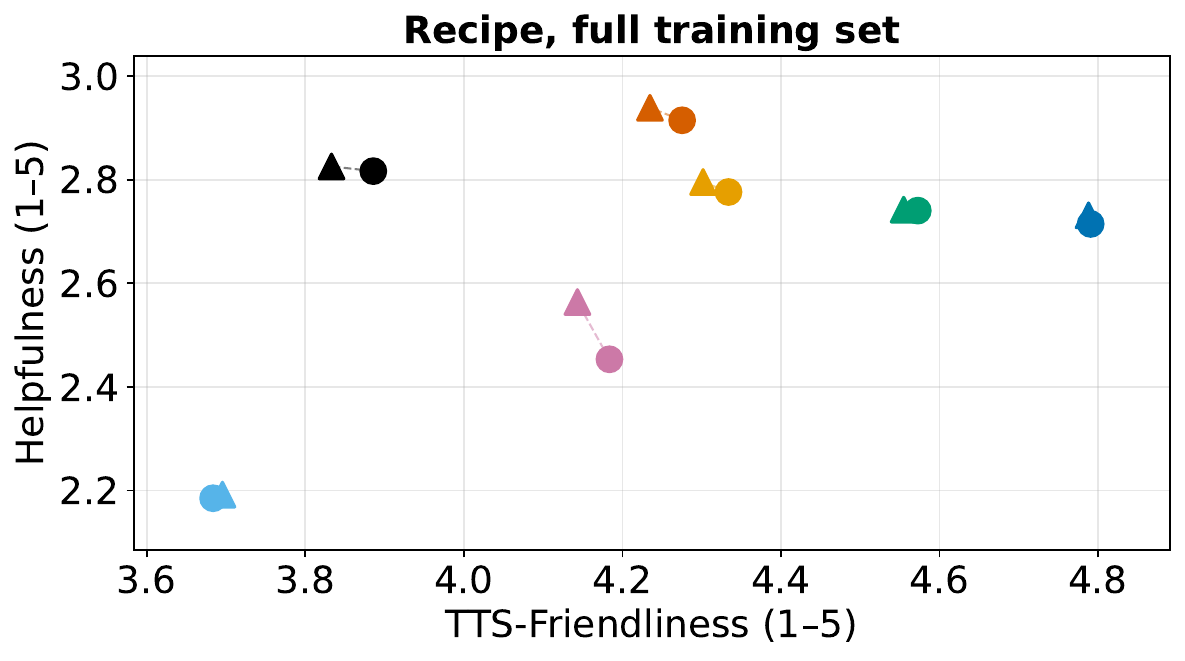}
    \end{subfigure}
    \hfill
    \begin{subfigure}[b]{0.49\textwidth}
        \includegraphics[width=\textwidth]{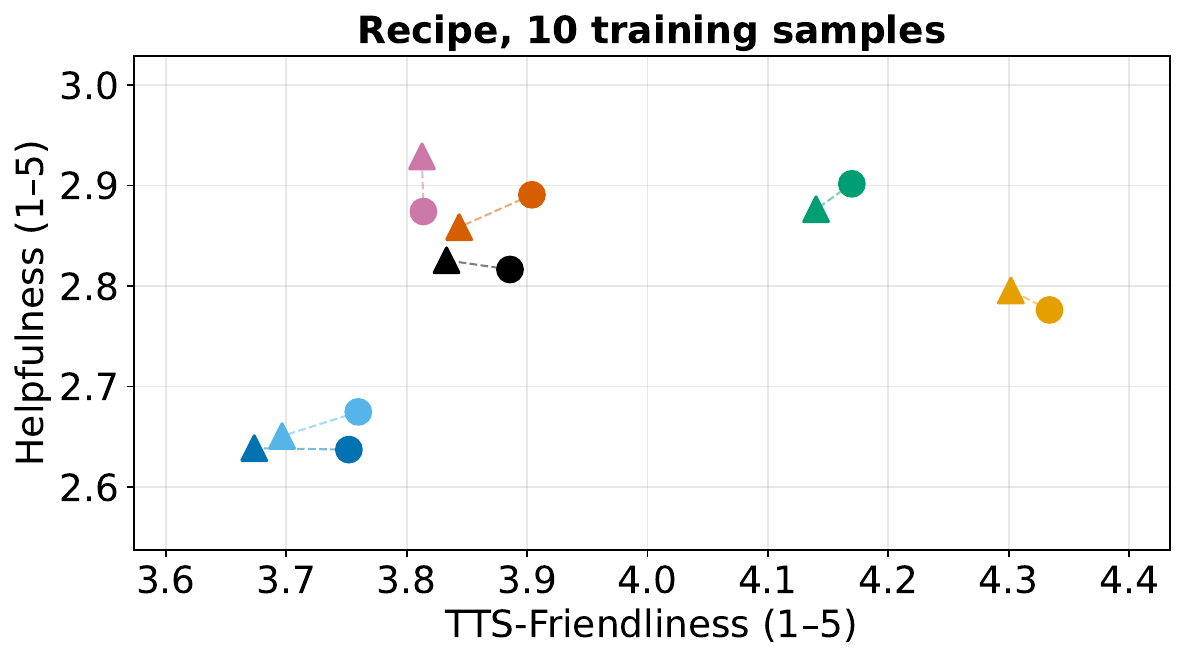}
    \end{subfigure}
    \caption{Comparison of generation approaches on the tradeoff between heuristic TTS-friendliness and Helpfulness (top-right is better). The base model used is Qwen3-4B. To improve readability, the x-axis and y-axis have been adjusted for each plot based on visible data points; they are not identical across the full-set and 10-sample settings.}
    \label{fig:results-qwen}
\end{figure*}

Next, we study the generation performance of all methods across the two datasets (CORA and Recipe) and under both training regimes (10-sample and full-data settings). Figure~\ref{fig:results-qwen} reports the results based on Qwen3-4B as the generation model, using the tradeoff between TTS-friendliness and helpfulness as the primary evaluation criterion. In complement, we provide results with another model family, SmolLM3-3B, in App.~\ref{app:results-smollm}.

\paragraph{FaST achieves the best overall tradeoff between TTS-friendliness and helpfulness.}

Across nearly all experimental conditions, FaST lies on or near the Pareto frontier, consistently achieving strong TTS-friendliness scores while maintaining high helpfulness. In contrast to methods that improve one objective at the expense of the other, FaST provides a more balanced score for both dimensions. This behavior is particularly visible on CORA, where FaST simultaneously improves TTS-friendliness and preserves helpfulness relative to Prompting and other preference-based baselines (DPO, GRPO-RM, RFT-RM). On Recipe, FaST still maintains competitive helpfulness while achieving among the highest TTS-friendliness scores (only beaten by SFT on the full setting and Prompting on the 10-sample setting).

\paragraph{FaST outperforms baselines based on the classical reward model.}

GRPO-RM and RFT-RM, which rely on a traditional reward model, are consistently outperformed by FaST across datasets and training regimes. In particular, FaST achieves higher TTS-friendliness while also preserving or improving helpfulness. These results suggest that relying on a feature-based reward model with very few learned parameters~-- as it is the case in FaST~-- provides an effective training signal for TTS-friendliness, unlike a traditional reward model with a large number of learned parameters. Interestingly, this observation nuances the preference prediction results (see Section~\ref{sec:response-prediction}), which showed that RM was always able to achieve high classification accuracy on validation and test. Therefore, while RM may be able to generalize on a classification task, the properties captured by its weights may not fully relate with those properties desirable for TTS-friendliness. In contrast, the features from FaRM explicitly capture such properties as we will describe in Section~\ref{sec:analysis}.

\paragraph{Prompting alone is insufficient for reliable TTS-friendly generation.}

Although prompting the model to generate TTS-friendly text provides some gains over the Zeroshot baseline, it underperforms FaST in most settings. The gap is especially pronounced on CORA, where FaST simultaneously achieve higher TTS-friendliness and Helpfulness. Therefore, while prompting may intuitively appear as a natural option for enforcing TTS-friendliness, the results suggest that it is difficult to achieve this property through prompting alone. Instead, we observe clear benefits from explicit alignment or fine-tuning procedures, and in particular from FaST and SFT.

\paragraph{FaST shines in low-data regimes; SFT suffices otherwise.}

The advantages of FaST are most pronounced when only a small number of training samples is available. In the 10-sample setting, FaST consistently outperforms other finetuning-based approaches (namely, SFT, DPO, GRPO-RM, and RFT-RM). This trend is especially clear on the Recipe dataset, where SFT exhibits a noticeable degradation in both dimensions while FaST remains comparatively stable. These results suggest that FaST is more sample-efficient and can leverage limited preference information more effectively than competing approaches. In contrast, when the full training set is available (100 training samples), standard supervised fine-tuning becomes highly competitive and provides a strong tradeoff between TTS-friendliness and Helpfulness. This goes in line with previous findings~\citep{LIMA} which showed that supervised fine-tuning from a few hundreds high-quality samples is often sufficient to obtain strong alignment performance.

\subsection{Analysis}
\label{sec:analysis}

\paragraph{Features discovered by FaST.} A key advantage of using FaST over traditional reward models is its interpretability. In our use-case, TTS-friendliness is decomposed into a set of explicit high-level features with independently learned weights. This provides direct insight into which properties are most associated with TTS-friendly generation in our datasets. Table~\ref{tab:cora-features10} presents the top-10 features discovered on CORA (using the full training set) and their learned weights. We also provide in App.~\ref{app:features} the full list of features discovered on CORA and Recipe (see Tables~\ref{tab:cora-features40} and~\ref{tab:recipe-features40}). The learned weights align closely with the intuitive principles of TTS-friendly text. Features associated with compact written format such as \textit{use\_of\_numeric\_formatting} ($-0.50$), \textit{abbreviation\_density} ($-0.32$), or \textit{technical\_code\_reference} ($-0.18$) receive strongly negative weights, supporting the intuition that symbolic notation and compressed forms are detrimental to spoken delivery. In contrast, conversational and speech-oriented properties such as \textit{natural\_conversational\_tone} ($+0.33$) and \textit{use\_of\_spelled\_out\_numbers} ($+0.33$) are positively rewarded, which is also aligned with expectations. Overall, the discovered features provide an interpretable characterization of the linguistic properties associated with improved TTS-friendliness on CORA's preference dataset.

\begin{table}[t]
    \centering
    \scalebox{0.9}{
    \begin{tabular}{lr}
    \toprule
    \textbf{Feature} & \textbf{Weight} \\
    \midrule
    use\_of\_numeric\_formatting & \colorvaluegradient{-0.50} \\
    natural\_conversational\_tone & \colorvaluegradient{0.33} \\
    use\_of\_spelled\_out\_numbers & \colorvaluegradient{0.33} \\
    abbreviation\_density & \colorvaluegradient{-0.32} \\
    external\_referral & \colorvaluegradient{-0.24} \\
    friendliness & \colorvaluegradient{0.23} \\
    descriptive\_sensory\_language & \colorvaluegradient{0.19} \\
    professional\_polish & \colorvaluegradient{0.18} \\
    technical\_code\_reference & \colorvaluegradient{-0.18} \\
    enthusiasm & \colorvaluegradient{0.17} \\
    \bottomrule
    \end{tabular}
    }
    \caption{Top-10 features discovered by FaST on CORA and their learned weights. Features are ordered by the magnitude of their weight in absolute value. Feature descriptions are provided in Table~\ref{tab:cora-features40}.}
    \label{tab:cora-features10}
\end{table}

\paragraph{Generated samples.}










\begin{table*}[t]
\centering
\small
\scalebox{0.97}{
\begin{tabular}{@{}p{1.7cm}p{13.3cm}@{}}
\toprule
\multicolumn{2}{@{}c@{}}{\textbf{CORA}} \\
\midrule

\multicolumn{2}{@{}p{15cm}@{}}{
\textbf{Context:} How much is the banana bread slice?
} \\[0.2cm]

\textbf{Responses:} & \\

\textit{Oracle:} &
The banana bread slice costs three dollars and forty-five cents. It's a delicious option with a brown butter drizzle that pairs wonderfully with our coffee selections. \\

\textit{Zeroshot:} &
The banana bread slice is \$3.45. \\

\textit{Prompting:} &
The banana bread slice costs \$3.45. \\

\textit{SFT:} &
The banana bread slice is three dollars and forty-five cents. It comes with a delightful brown butter drizzle. \\

\textit{DPO:} &
The banana bread slice is \$3.45 and comes with a brown butter drizzle. It's a perfect pairing for a morning or afternoon treat! \\

\textit{GRPO-RM:} &
The Banana Bread Slice with Brown Butter Drizzle is priced at \$3.45. I'll place your order for the Banana Bread Slice with Brown Butter Drizzle. Your coffee will be ready for you at the coffeehouse. \\

\textit{RFT-RM:} &
The banana bread slice is \$3.45. \\

\textit{FaST:} &
A banana bread slice with a brown butter drizzle is three point forty-five dollars. \\

\bottomrule
\end{tabular}
}
\caption{Example generations from the different approaches on the CORA dataset.}
\label{tab:cora-samples-all}
\end{table*}

Generated samples further illustrate the qualitative differences between the different approaches. As shown in Table~\ref{tab:cora-samples-all} for CORA, FaST and SFT tend to produce responses that are more natural to read aloud, favoring spelled-out numbers, conversational phrasing, while avoiding dense symbolic or abbreviated formats that are less suitable for TTS. In contrast, other methods often retain written-style conventions such as compact numeric expressions, which can negatively impact spoken delivery despite remaining factually correct. We provide additional generated samples in App.~\ref{app:generations} (see Table~\ref{tab:recipe-samples-all}). 

\subsection{Comparison against Text Normalization}
\label{sec:text-normalization}

As motivated in Section~\ref{sec:introduction}, our work focuses on direct TTS-friendly generation instead of a two-step procedure consisting of generating a first TTS-unfriendly response, and then rewriting it using a text normalization approach. To provide a comparison of these two paradigms, we experimented with PolyNorm~\citep{Wong2025tn}, a few-shot LLM-based text normalization method. We adopted Qwen3-4B as the prompted model and applied PolyNorm to the Zeroshot output to normalize it. We report heuristic TTS-friendliness, Helpfulness and Latency for Zeroshot, PolyNorm and FaST in Table~\ref{tab:text-normalization}. These results confirm that a dedicated text normalization stage can greatly improve the TTS-friendliness of the original Zeroshot response. This however comes at the cost of a doubled inference time due to the two required generation steps~-- the initial response generation and the rewriting step. In comparison, FaST achieves a highly competitive TTS-friendliness score in a single inference step, significantly reducing latency. 


\begin{table*}[t]
\centering
\scalebox{0.87}{
\begin{tabular}{@{}lrrrrrr@{}}
\toprule
\multirow{2}{*}{\textbf{Approach}} & \multicolumn{3}{c}{\textbf{CORA}} & \multicolumn{3}{c}{\textbf{Recipe}} \\
\cmidrule(lr){2-4} \cmidrule(ll){5-7}
 & \textbf{TTS-friendliness} $\uparrow$  & \textbf{Helpfulness} $\uparrow$ & \textbf{Latency} $\downarrow$ & \textbf{TTS-friendliness} $\uparrow$ & \textbf{Helpfulness} $\uparrow$ & \textbf{Latency} $\downarrow$ \\
\midrule
Zeroshot & 3.84 & 4.72 & 1.6 & 3.86 & 2.82 & 4.4 \\
PolyNorm & 4.92 & 4.72 & 3.4 & 4.88 & 2.75 & 8.5 \\
FaST & 4.73 & 4.86 & 1.6 & 4.56 & 2.74 & 4.4 \\
\bottomrule
\end{tabular}
}
\caption{Comparison of the heuristic TTS-friendliness, Helpfulness, and Latency of FaST, the text normalization baseline PolyNorm, and Zeroshot on CORA and Recipe. Latency corresponds to the average generation time for a single response, in seconds. Higher is better for TTS-friendliness and Helpfulness; lower is better for Latency.}
\label{tab:text-normalization}
\end{table*}

\section{Validating our Heuristic Metric as a Reliable TTS-friendliness Proxy}
\label{sec:validation}

Throughout our experiments, we rely on the Heuristic metric as the primary 
evaluation signal given its low cost. We now validate this choice by showing it 
is a reliable proxy for the more expensive TTS$\to$ASR pipeline and human judgments.

\subsection{Comparison with the TTS$\to$ASR Metric}
\label{sec:correlations}

We compare the Heuristic and TTS$\to$ASR metrics across the outputs of five representative approaches on both CORA and Recipe.

\paragraph{Protocol.} For each dataset, we score the test utterances of the first split produced by five systems (Oracle, Zeroshot, Prompting, DPO, FaST), yielding $5 \times 81 = 405$ rated utterances on CORA and $5 \times 100 = 500$ on Recipe. We then compute, separately for each dataset, the Spearman rank correlation between the Heuristic score and each TTS$\to$ASR metric (CER, WER), both at the utterance level (pooled across systems) and at the system level (correlation of per-system means, $n=5$).

\begin{table}[t]
  \centering
  \scalebox{0.79}{
  \begin{tabular}{@{}lrrrrrr@{}}
    \toprule
    \multirow{2}{*}{\textbf{Approach}} & \multicolumn{3}{c}{\textbf{CORA}} & \multicolumn{3}{c}{\textbf{Recipe}} \\
    \cmidrule(lr){2-4} \cmidrule(ll){5-7}
    & \textbf{Heur $\uparrow$} & \textbf{CER $\downarrow$} & \textbf{WER $\downarrow$} & \textbf{Heur $\uparrow$} & \textbf{CER $\downarrow$} & \textbf{WER $\downarrow$} \\
    \midrule
    Oracle & 4.838 & 0.051 & 0.260 & 4.799 & 0.044 & 0.228 \\
    FaST & 4.648 & 0.077 & 0.297 & 4.512 & 0.066 & 0.253 \\
    DPO & 4.010 & 0.141 & 0.377 & 4.079 & 0.117 & 0.297 \\
    Prompt. & 3.999 & 0.206 & 0.478 & 4.324 & 0.123 & 0.349 \\
    Zeroshot & 3.836 & 0.228 & 0.523 & 3.763 & 0.170 & 0.420 \\
    \bottomrule
  \end{tabular}
  }
 \caption{Per-model means sorted by CORA CER ($n=81$ per model on CORA, $n=100$ on Recipe). Heur: higher is better; CER/WER: lower is better. 
}
  \label{tab:corr_combined}
\end{table}

\paragraph{Discussion.} The Heuristic and TTS$\to$ASR metrics produce closely matching system rankings in Table~\ref{tab:corr_combined}: identical on CORA (Oracle > FaST > DPO > Prompting > Zeroshot, yielding a perfect system-level Spearman $\rho=-1.00$) and differing only by a single DPO/Prompting swap on Recipe ($\rho=-0.90$). The utterance-level pooled Spearman correlations are strong and negative on both datasets: $\rho_{\text{Heur-CER}}=-0.71$ and $\rho_{\text{Heur-WER}}=-0.68$ on CORA, $n=405$; $\rho_{\text{Heur-CER}}=-0.80$ and $\rho_{\text{Heur-WER}}=-0.71$ on Recipe, $n=500$ (all $p\ll 0.001$).\footnote{The negative sign is expected: higher Heuristic scores predict lower TTS$\to$ASR error rates.} We conclude that the Heuristic score is a cheap and reliable proxy for the empirical TTS$\to$ASR metric on both datasets.

\subsection{Validation via Human Judges}


We complement the empirical TTS$\to$ASR analysis with a MUSHRA-style~\cite{itu_bs1534}
listening test conducted on the Prolific crowdsourcing platform.\footnote{\url{https://www.prolific.com}}
We adapted the protocol to our setting: standard MUSHRA evaluates audio encoding
or synthesis quality, whereas we hold the TTS engine and the voice
constant across systems and ask participants to judge the
\emph{listenability of the spoken content itself} (i.e., how naturally the 
response is rendered when read aloud). 
The full protocol, hosting setup, and
quality-control rules are detailed in App.~\ref{app:mushra_details}.

From the CORA test split we sampled 20 questions and synthesized each
system's response with Kyutai Pocket TTS using a single fixed voice. 
For each question, participants rated the  Oracle reference and three
 shuffled generated answers (Prompting, DPO, FaST) on a
0--100 slider.\footnote{Only 3 systems were included in the listening test to limit the cognitive load for judges and fit within an approximate budget of \pounds300 for Prolific  crowdsourced human evaluation.} The study was carried out in two Prolific batches and reached
$N=14$ valid raters after quality control.

\begin{figure}[t]
  \centering
  \includegraphics[width=0.97\linewidth]{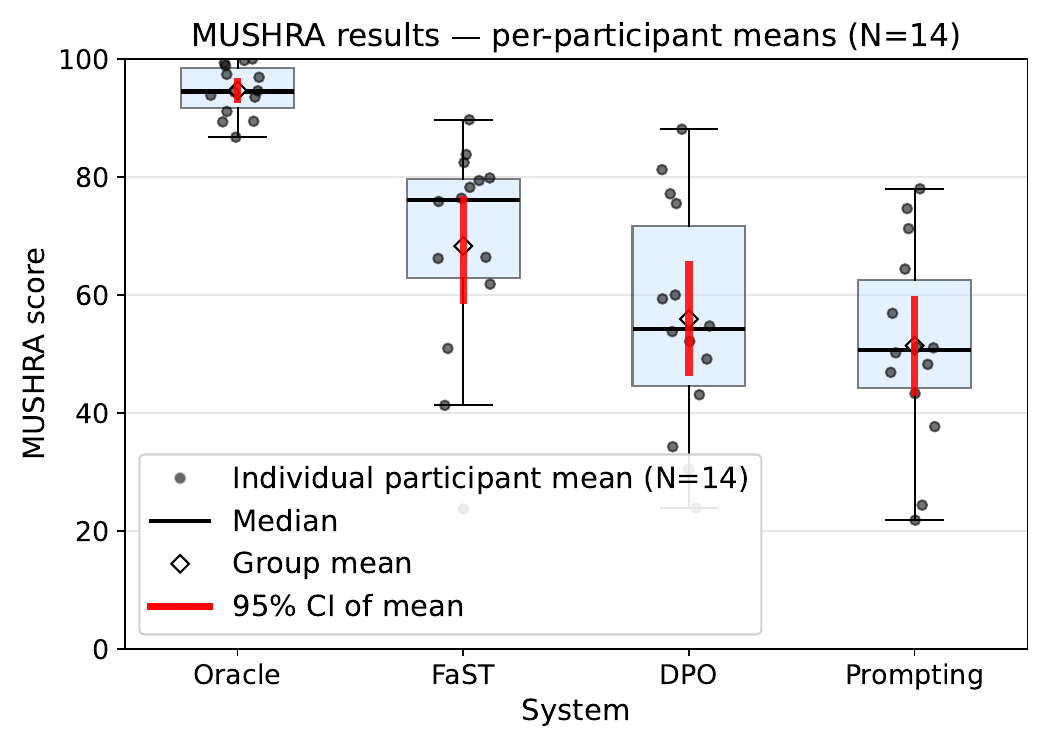}
  \caption{MUSHRA listening test: per-participant mean rating per system.
  Boxes show the median and inter-quartile range across participants; black
  dots are individual participants; the white diamond marks the group mean;
  the red bar is the 95\% bootstrap confidence interval of the mean.
  Sample size: $N=14$ human judges.}
  \label{fig:mushra_boxplot}
\end{figure}

Figure~\ref{fig:mushra_boxplot} reports, for each system, the distribution of
per-participant mean ratings across the 20 trials. The Oracle is rated
near the top of the scale (mean 94.7, 95\% CI $[92.5, 96.8]$), confirming
that participants used the slider as intended. The ranking obtained on the
trained systems is consistent with the automatic metrics reported in
Section~\ref{sec:correlations}: FaST is rated highest
(68.3, $[58.5, 76.8]$), ahead of DPO (55.9, $[46.3, 65.8]$) and
Prompting (51.4, $[42.8, 59.9]$). We confirm pairwise contrasts
between compared systems with paired Wilcoxon signed-rank tests on the
per-participant means:
\emph{FaST}~$>$~\emph{DPO} (mean difference of $+12.4$, $p=0.003$),
\emph{FaST}~$>$~\emph{Prompting} ($+16.9$, $p=0.001$).
App.~\ref{app:mushra_details} further reports a strong correlation between the heuristic
metric and human MUSHRA ratings.

\section{Conclusion}
\label{sec:conclusion}

We introduced the problem of TTS-friendly text generation through LLM alignment, motivated by the observation that current language models are optimized primarily for written text rather than spoken delivery. To support research in this setting, we proposed two preference datasets~-- CORA and Recipe~-- spanning the coffee ordering and cooking domains, together with an evaluation suite combining a heuristic metric, TTS$\to$ASR evaluation, and a study with human judges. Our experiments across datasets and settings demonstrated that the recent feature-based FaST approach overall achieves the best tradeoff between TTS-friendliness and helpfulness, with particularly strong gains when only a handful of training samples are available. Additionally, we showed that our heuristic metric is strongly correlated with the TTS$\to$ASR metric and human judgments, which suggests that TTS-friendliness can be efficiently assessed using pattern-based heuristics. This opens up applications where such metric could be directly utilized as a reward model, similarly to recent work on reinforcement learning from verifiable rewards~\citep{rlvr}. More broadly, our results highlight the importance of optimizing language generation not only for readability and helpfulness, but also for downstream spoken interaction, paving the way toward more natural and accessible voice-based conversational assistants.

\section*{Limitations}

\paragraph{Response length.}
FaST overall tends to generate longer responses than other approaches, as evidenced by the average response lengths shown in Table~\ref{tab:lengths} (see App.~\ref{app:length}). Given that TTS-unfriendly responses are typically 
shorter and more compact while TTS-friendly responses tend to be more verbose, FaST 
has learned to associate response length with TTS-friendliness. This is reflected in 
Tables~\ref{tab:cora-features40} and~\ref{tab:recipe-features40} which report the features discovered by FaST on CORA and Recipe, respectively: the features \textit{conciseness} and \textit{brevity} carry negative weights (respectively, $-0.12$ and $-0.25$). While longer responses are not inherently bad for TTS, 
this bias toward verbosity may not be desirable in all applications. This could be easily addressed in practice by manually setting the weight of the \textit{conciseness} or \textit{brevity} feature to zero~-- or even to a small positive value to further penalize length. App.~\ref{app:verbosity-control} details an experiment which supports the validity of this fix.

\paragraph{Dataset, domain, and language coverage.}
As no established benchmark for TTS-friendly text generation exists, we had to 
construct both datasets from scratch, which required non-trivial effort even with 
synthetic generation pipelines. Our experiments are further limited to English and 
two domains; some TTS-unfriendly patterns may vary across languages and domains 
such as medical, legal, or technical content. Extending our approach to a broader 
range of languages and domains would require building additional preference data 
to improve out-of-the-box applicability.

\paragraph{Generalization  to larger LLMs.}
Our empirical study is constrained to 3B/4B-parameter models, which represent the largest configurations that our computational budget permitted us to fine-tune on a single GPU within the time requirements imposed by the alignment procedures. Nonetheless, our evaluation encompasses two distinct model families (Qwen3-4B and SmolLM3-3B), and FaST exhibits consistently competitive performance across both, indicating that the observed effect is unlikely to be model-specific. Whether this behavior persists at larger model scales remains an open question.

\paragraph{Generalization  to different TTS systems.}
Our MUSHRA evaluation relies on a single TTS engine, which limits the strength of our claim regarding engine independence. Our objective, however, is to assess and improve the intrinsic TTS-compatibility of the LLM-generated text itself, independently of any particular downstream synthesizer. This focus is motivated by the substantial variability in normalization front-ends across TTS systems, as well as the tendency of lightweight engines to employ considerably simpler preprocessing pipelines. Rather than tuning to a specific system, we aim to generate text that is easier to synthesize for a broad range of TTS engines. This decoupling between text generation and synthesis is particularly critical in settings where only compact models are available, such as embedded TTS in robotics.

\paragraph{Cascade architecture assumption.}
Our work targets cascade LLM$\to$TTS architectures~-- the dominant paradigm for 
voice-based dialogue systems. End-to-end speech LLMs that directly generate speech 
tokens are an emerging alternative to which our framework does not directly apply. 
We believe however that aligning spoken-delivery quality via interpretable 
text-level features could be adapted to guide intermediate representations in speech 
LLMs, and leave this extension to future work.
\section*{Ethical Considerations}
\label{sec:ethics}

\paragraph{Human listening study.} We ran a MUSHRA listening study on
Prolific with $N=14$ participants who passed our quality-control checks.
Participants gave informed consent via the Prolific study listing (which
declared the task as academic research) and via the platform's
Participant Terms accepted at registration. Each participant was paid
\pounds 12 for an estimated 40-minute task ($\approx$ \pounds 18/h),
above both the UK National Living Wage and Prolific's recommended
minimum; participants whose submissions failed our post-hoc
quality-control checks were nevertheless paid in full. No personally
identifying information, demographic data, or audio recording from the
participant were collected; we recorded only the per-trial slider ratings,
per-trial timestamps, the anonymous Prolific ID (used only to release
payment, never redistributed), and an optional free-text comment. The
task~-- listening to short synthetic speech clips and moving 0--100
sliders~-- involves no sensitive content, no deception, no biometric
data, and no personal information; we judged that it does not fall
within the scope of mandatory ethics review. Full instructions,
screenshots of the interface, and detailed quality-control criteria are
provided in App.~\ref{app:mushra_details}.

\paragraph{Dataset.} The conversational stimuli are taken from CORA, a
synthetic customer-service dataset created for this paper. CORA contains
no real user data, no personally identifying information, and no
sensitive content.

\paragraph{External LLM judge.} One of our scoring methods relies on a
black-box call to OpenAI's GPT-4o-mini API to rate the helpfulness of
generated text. Only the model-generated assistant responses are sent
to the API; no participant data, no human-written text from the dataset
beyond the released synthetic stimuli, and no other identifying
information are transmitted. We treat the API as a scoring service and
do not fine-tune or train on its outputs.

\paragraph{TTS engine and dual use.} The audio stimuli are synthesised
with Kyutai Pocket TTS, a publicly released model, using a single fixed
voice (\textit{alba}) for all systems and all utterances. No
individual's voice is cloned or imitated. Making AI-generated responses
more natural when spoken aloud is intended for voice-assistant and
accessibility use cases; we do not see a specific misuse risk beyond
that already inherent to publicly available TTS systems.

\paragraph{Use of AI assistants.} Beyond their role as LLM judges and
dataset construction tools described above, AI assistants were used
during the preparation of this work to support code writing and the
drafting of certain passages of this paper. In all cases, AI-generated
suggestions were reviewed, edited, and validated by the authors; no
content was incorporated without human oversight, and the assistants
were never used in a fully automated fashion. The authors take full
responsibility for the content of this paper.

{
    \small
    \bibliographystyle{ieeenat_fullname}
    \bibliography{main}

@inproceedings{speechworthy2024,
  author    = {Hyundong Justin Cho and Nicolaas Paul Jedema and Leonardo F. R. Ribeiro 
               and Karishma Sharma and Pedro Szekely and Alessandro Moschitti 
               and Ruben Janssen and Jonathan May},
  title     = {Speechworthy Instruction-tuned Language Models},
  booktitle = {Proceedings of the 2024 Conference on Empirical Methods in Natural 
               Language Processing},
  pages     = {10652--10670},
  year      = {2024},
  doi       = {10.18653/v1/2024.emnlp-main.595},
  url       = {https://aclanthology.org/2024.emnlp-main.595/}
}

@inproceedings{xie2024controllable,
  author    = {Tianxin Xie and Yan Rong and Pengfei Zhang and Wenwu Wang and Li Liu},
  title     = {Towards Controllable Speech Synthesis in the Era of Large Language 
               Models: A Systematic Survey},
  booktitle = {Proceedings of the 2025 Conference on Empirical Methods in Natural 
               Language Processing},
  pages     = {764--791},
  year      = {2025},
  doi       = {10.18653/v1/2025.emnlp-main.40},
  url       = {https://aclanthology.org/2025.emnlp-main.40/}
}

@inproceedings{zhang2025advancing,
  author    = {Zhang, Xueyao and Wang, Yuancheng and Wang, Chaoren and
               Li, Ziniu and Chen, Zhuo and Wu, Zhizheng},
  title     = {Advancing Zero-shot Text-to-Speech Intelligibility across Diverse 
               Domains via Preference Alignment},
  booktitle = {Proceedings of the 63rd Annual Meeting of the Association for 
               Computational Linguistics (Volume 1: Long Papers)},
  pages     = {12251--12270},
  year      = {2025},
  doi       = {10.18653/v1/2025.acl-long.598},
  url       = {https://aclanthology.org/2025.acl-long.598/}
}

@misc{mathreader2024,
  author        = {Sieun Hyeon and Kyudan Jung and Nam-Joon Kim and 
                   Hyun Gon Ryu and Jaeyoung Do},
  title         = {{MathReader}: Text-to-Speech for Mathematical Documents},
  year          = {2025},
  eprint        = {2501.07088},
  archivePrefix = {arXiv},
  primaryClass  = {cs.CL}
}

@inproceedings{tian2025preference,
  author    = {Jinchuan Tian and Chunlei Zhang and Jiatong Shi and Hao Zhang and
               Jianwei Yu and Shinji Watanabe and Dong Yu},
  title     = {Preference Alignment Improves Language Model-Based {TTS}},
  booktitle = {Proceedings of the 2025 IEEE International Conference on Acoustics, 
               Speech and Signal Processing},
  pages     = {1--5},
  year      = {2025},
  url       = {https://ieeexplore.ieee.org/abstract/document/10890510}
}

@inproceedings{samuel2025cie,
  author    = {Samuel, Vinay and Diddee, Harshita and Zhang, Yiming and 
               Ippolito, Daphne},
  title     = {{CIE}: Controlling Language Model Text Generations Using 
               Continuous Signals},
  booktitle = {Proceedings of the 2025 Conference on Empirical Methods in 
               Natural Language Processing},
  pages     = {3815--3825},
  year      = {2025},
  doi       = {10.18653/v1/2025.emnlp-main.189},
  url       = {https://aclanthology.org/2025.emnlp-main.189/}
}

@inproceedings{hussain2025koel,
  author    = {Hussain, Shehzeen Samarah and Neekhara, Paarth and Yang, Xuesong and
               Casanova, Edresson and Ghosh, Subhankar and Fejgin, Roy and
               Desta, Mikyas T. and Valle, Rafael and Li, Jason},
  title     = {Koel-{TTS}: Enhancing {LLM} based Speech Generation with Preference 
               Alignment and Classifier Free Guidance},
  booktitle = {Proceedings of the 2025 Conference on Empirical Methods in 
               Natural Language Processing},
  pages     = {21219--21234},
  year      = {2025},
  doi       = {10.18653/v1/2025.emnlp-main.1076},
  url       = {https://aclanthology.org/2025.emnlp-main.1076/}
}

@inproceedings{xie2025shotbyshot,
  author    = {Xie, Junyu and Han, Tengda and Bain, Max and Nagrani, Arsha and
               Khandelwal, Eshika and Varol, G{\"u}l and Xie, Weidi and 
               Zisserman, Andrew},
  title     = {Shot-by-Shot: Film-Grammar-Aware Training-Free Audio Description 
               Generation},
  booktitle = {Proceedings of the 2025 IEEE/CVF International Conference on 
               Computer Vision},
  pages     = {16503--16513},
  month     = {October},
  year      = {2025},
  url       = {https://arxiv.org/abs/2504.01020}
}

@inproceedings{cardei2025constrained,
  author    = {Cardei, Michael and Christopher, Jacob K. and Hartvigsen, Thomas and
               Bartoldson, Brian R. and Kailkhura, Bhavya and Fioretto, Ferdinando},
  title     = {Constrained Discrete Diffusion},
  booktitle = {Proceedings of the 39th Conference on Neural Information Processing Systems},
  year      = {2025},
  url       = {https://arxiv.org/abs/2503.09790}
}

@misc{duran2025beyond,
  author        = {Mehmet Samet Duran and Tevfik Aytekin},
  title         = {Beyond One-Size-Fits-All Summarization: Customizing Summaries 
                   for Diverse Users},
  year          = {2025},
  eprint        = {2503.10675},
  archivePrefix = {arXiv},
  primaryClass  = {cs.CL}
}

@inproceedings{hirsch2021clid,
  author    = {Elad Hirsch and Ayellet Tal},
  title     = {{CLID}: Controlled-Length Image Descriptions with Limited Data},
  booktitle = {Proceedings of the 2024 IEEE/CVF Winter Conference on Applications 
               of Computer Vision},
  year      = {2024},
  eprint    = {2211.14835},
  archivePrefix = {arXiv},
  primaryClass  = {cs.CV}
}

@techreport{itu_bs1534,
  author      = {{International Telecommunication Union}},
  title       = {{Method for the Subjective Assessment of Intermediate Quality 
                 Level of Audio Systems}},
  type        = {Recommendation},
  number      = {ITU-R BS.1534-3},
  institution = {International Telecommunication Union, Radiocommunication 
                 Sector (ITU-R)},
  year        = {2015},
  month       = oct,
  url         = {https://www.itu.int/rec/R-REC-BS.1534-3-201510-I}
}

@article{schoeffler2018webmushra,
  author    = {Schoeffler, Michael and Bartoschek, Sarah and 
               St{\"o}ter, Fabian-Robert and Ro{\ss}, Marlene and 
               Westphal, Susanne and Edler, Bernd and Herre, J{\"u}rgen},
  title     = {{webMUSHRA} --- {A} Comprehensive Framework for Web-based 
               Listening Tests},
  journal   = {Journal of Open Research Software},
  volume    = {6},
  number    = {1},
  pages     = {8},
  year      = {2018},
  publisher = {Ubiquity Press},
  doi       = {10.5334/jors.187}
}

@inproceedings{bien2020recipenlg,
  author    = {Bie{\'n}, Micha{\l} and Gilski, Micha{\l} and 
               Maciejewska, Martyna and Taisner, Wojciech and 
               Wisniewski, Dawid and Lawrynowicz, Agnieszka},
  title     = {{RecipeNLG}: A Cooking Recipes Dataset for Semi-Structured 
               Text Generation},
  booktitle = {Proceedings of the 13th International Conference on Natural 
               Language Generation},
  pages     = {22--28},
  year      = {2020},
  url       = {https://aclanthology.org/2020.inlg-1.4/}
}

@misc{kyutai_pockettts,
  author       = {{Kyutai Labs}},
  title        = {{Pocket TTS}: A Compact {CPU}-Friendly Streaming 
                 Text-to-Speech Model},
  year         = {2025},
  howpublished = {\url{https://github.com/kyutai-labs/pocket-tts}},
  note         = {Software release}
}

@misc{wav2vec2_large_960h_lv60_self,
  author       = {{Facebook AI Research}},
  title        = {{wav2vec2-large-960h-lv60-self}},
  year         = {2021},
  howpublished = {\url{https://huggingface.co/facebook/wav2vec2-large-960h-lv60-self}},
  note         = {HuggingFace model card}
}

@inproceedings{fast,
  author    = {Thonet, Thibaut and Kruszewski, Germ{\'a}n and Rozen, Jos 
               and Erbacher, Pierre and Dymetman, Marc},
  title     = {{F}a{ST}: Feature-aware Sampling and Tuning for Personalized 
               Preference Alignment with Limited Data},
  booktitle = {Proceedings of the 2025 Conference on Empirical Methods in 
               Natural Language Processing},
  pages     = {9341--9370},
  year      = {2025},
  doi       = {10.18653/v1/2025.emnlp-main.475},
  url       = {https://aclanthology.org/2025.emnlp-main.475/}
}

@inproceedings{
lima,
title={{LIMA}: Less Is More for Alignment},
author={Chunting Zhou and Pengfei Liu and Puxin Xu and Srini Iyer and Jiao Sun and Yuning Mao and Xuezhe Ma and Avia Efrat and Ping Yu and Lili Yu and Susan Zhang and Gargi Ghosh and Mike Lewis and Luke Zettlemoyer and Omer Levy},
booktitle={Proceedings of the 37th Conference on Neural Information Processing Systems},
year={2023},
url={https://openreview.net/forum?id=KBMOKmX2he}
}

@inproceedings{
dpo,
title={Direct Preference Optimization: Your Language Model is Secretly a Reward Model},
author={Rafael Rafailov and Archit Sharma and Eric Mitchell and Christopher D. Manning and Stefano Ermon and Chelsea Finn},
booktitle={Proceedings of the 37th Conference on Neural Information Processing Systems},
year={2023},
url={https://openreview.net/forum?id=HPuSIXJaa9}
}

@misc{grpo,
      title={{DeepSeekMath}: Pushing the Limits of Mathematical Reasoning in Open Language Models}, 
      author={Zhihong Shao and Peiyi Wang and Qihao Zhu and Runxin Xu and Junxiao Song and Xiao Bi and Haowei Zhang and Mingchuan Zhang and Y. K. Li and Y. Wu and Daya Guo},
      year={2024},
      eprint={2402.03300},
      archivePrefix={arXiv},
      primaryClass={cs.CL},
      url={https://arxiv.org/abs/2402.03300}, 
}

@inproceedings{
rlvr,
title={Reinforcement Learning with Verifiable Rewards Implicitly Incentivizes Correct Reasoning in Base {LLM}s},
author={Xumeng Wen and Zihan Liu and Shun Zheng and Shengyu Ye and Zhirong Wu and Yang Wang and Zhijian Xu and Xiao Liang and Junjie Li and Ziming Miao and Jiang Bian and Mao Yang},
booktitle={Proceedings of the 14th International Conference on Learning Representations},
year={2026},
url={https://openreview.net/forum?id=jGbRWwIidy}
}

@INPROCEEDINGS{Zhang2024tn,
  author={Zhang, Yang and Bartley, Travis M. and Graterol-Fuenmayor, Mariana and Lavrukhin, Vitaly and Bakhturina, Evelina and Ginsburg, Boris},
  booktitle={Proceedings of the 2024 IEEE International Conference on Acoustics, 
               Speech and Signal Processing}, 
  title={A Chat about Boring Problems: Studying GPT-Based Text Normalization}, 
  year={2024},
  volume={},
  number={},
  pages={10921-10925},
  doi={10.1109/ICASSP48485.2024.10447169}}

@inproceedings{Wong2025tn,
  author       = {Michel Wong and
                  Ali Alshehri and
                  Sophia Kao and
                  Haotian He},
  editor       = {Saloni Potdar and
                  Lina Maria Rojas{-}Barahona and
                  S{\'{e}}bastien Montella},
  title        = {{PolyNorm}: Few-Shot LLM-Based Text Normalization for Text-to-Speech},
  booktitle    = {Proceedings of the 2025 Conference on Empirical Methods in Natural
                  Language Processing (Industry Track)},
  pages        = {77--85},
  year         = {2025},
  url          = {https://doi.org/10.18653/v1/2025.emnlp-industry.6},
  doi          = {10.18653/V1/2025.EMNLP-INDUSTRY.6},
  bibsource    = {dblp computer science bibliography, https://dblp.org}
}

@inproceedings{kim-etal-2024-prometheus-1,
    title = "Prometheus: Inducing Fine-grained Evaluation Capability in Language Models",
    author = "Kim, Seungone and Shin, Jamin and Cho, Yejin and Jang, Joel and
      Longpre, Shayne and Lee, Hwaran and Yun, Sangdoo and Shin, Seongjin and
      Kim, Sungdong and Thorne, James and Seo, Minjoon",
    booktitle = "Proceedings of the 12th International Conference on Learning Representations",
    year = "2024",
    url = "https://openreview.net/forum?id=8euJaTveKw"
}

@inproceedings{kim-etal-2024-prometheus,
    title = "Prometheus 2: An Open Source Language Model Specialized in Evaluating Other Language Models",
    author = "Kim, Seungone  and
      Suk, Juyoung  and
      Longpre, Shayne  and
      Lin, Bill Yuchen  and
      Shin, Jamin  and
      Welleck, Sean  and
      Neubig, Graham  and
      Lee, Moontae  and
      Lee, Kyungjae  and
      Seo, Minjoon",
    editor = "Al-Onaizan, Yaser  and
      Bansal, Mohit  and
      Chen, Yun-Nung",
    booktitle = "Proceedings of the 2024 Conference on Empirical Methods in Natural Language Processing",
    month = nov,
    year = "2024",
    url = "https://aclanthology.org/2024.emnlp-main.248/",
    doi = "10.18653/v1/2024.emnlp-main.248",
    pages = "4334--4353"
}

@inproceedings{thonet-etal-2025-elitr,
    title = "{ELITR-Bench}: A Meeting Assistant Benchmark for Long-Context Language Models",
    author = "Thonet, Thibaut  and
      Besacier, Laurent  and
      Rozen, Jos",
    booktitle = "Proceedings of the 31st International Conference on Computational Linguistics",
    year = "2025",
    url = "https://aclanthology.org/2025.coling-main.28/",
    pages = "407--428"
}
}

\clearpage
\appendix

\section{Details on MUSHRA Tests with Prolific}
\label{app:mushra_details}

\paragraph{Stimuli.} We sampled 20 candidate questions from CORA's test
set (from the first split, which contains a total of 81 unique questions in the test set). 
For each question, four responses were synthesized with Kyutai Pocket
TTS\footnote{Kyutai Pocket TTS, 100M parameters; voice \textit{alba}; 24~kHz
mono; CPU inference.} using a single fixed voice: the \textit{chosen} response
(\texttt{Oracle}, used as the labeled reference) and the outputs of three
systems (Prompting, DPO, FaST). To comply with
the webMUSHRA equal-duration requirement, each set of four clips was
zero-padded with trailing silence to match the longest clip in the set.

\paragraph{Protocol.} The study was implemented with the
webMUSHRA framework~\citep{schoeffler2018webmushra}. Each trial presented one question, the labeled reference (Oracle), and four anonymous, shuffled
conditions (Prompting, DPO, FaST, and Oracle again for quality control); the participant
rated each clip on a 0--100 slider.

We adapted the ITU-R~BS.1534 MUSHRA protocol~\cite{itu_bs1534} to our setting
because our evaluation target differs from the original methodology. The
standard was designed to assess the quality of audio coding or synthesis
(different encoders/synthesizers applied to the same source), whereas in our
experiment the synthesizer and voice are constant across systems and what
differs is the \emph{text} produced by each system. The participants are
therefore asked to rate how naturally and intelligibly the spoken response
answers the question  (i.e. the listenability of the textual
response when rendered as speech) rather than the perceived quality of the
audio itself. Our main deviation from the standard is the removal of the 3.5~kHz low-pass anchor, which is irrelevant here: our independent variable is the spoken message, not audio encoding quality.

All other MUSHRA conventions are preserved. In particular, each trial presents
the Oracle reference both as a labeled clip at the top left of the page (for
explicit comparison) and as a hidden anonymous copy mixed with the three
system outputs, yielding four shuffled anonymous conditions to be
rated on 0--100 sliders (Prompting, DPO, FaST, and Oracle). The hidden reference doubles as a built-in attention check:
participants' contributions whose ratings of this clip fall below~80 on more than 20\% of
trials are excluded from our study. The remaining conventions
(continuous slider, simultaneous playback of all conditions, equal duration
via silence-padding, and per-trial shuffling) are unchanged.
The study was hosted on an inhouse server
running Apache and a small Python
backend.

\begin{figure*}[t]
  \centering
  \includegraphics[width=\linewidth]{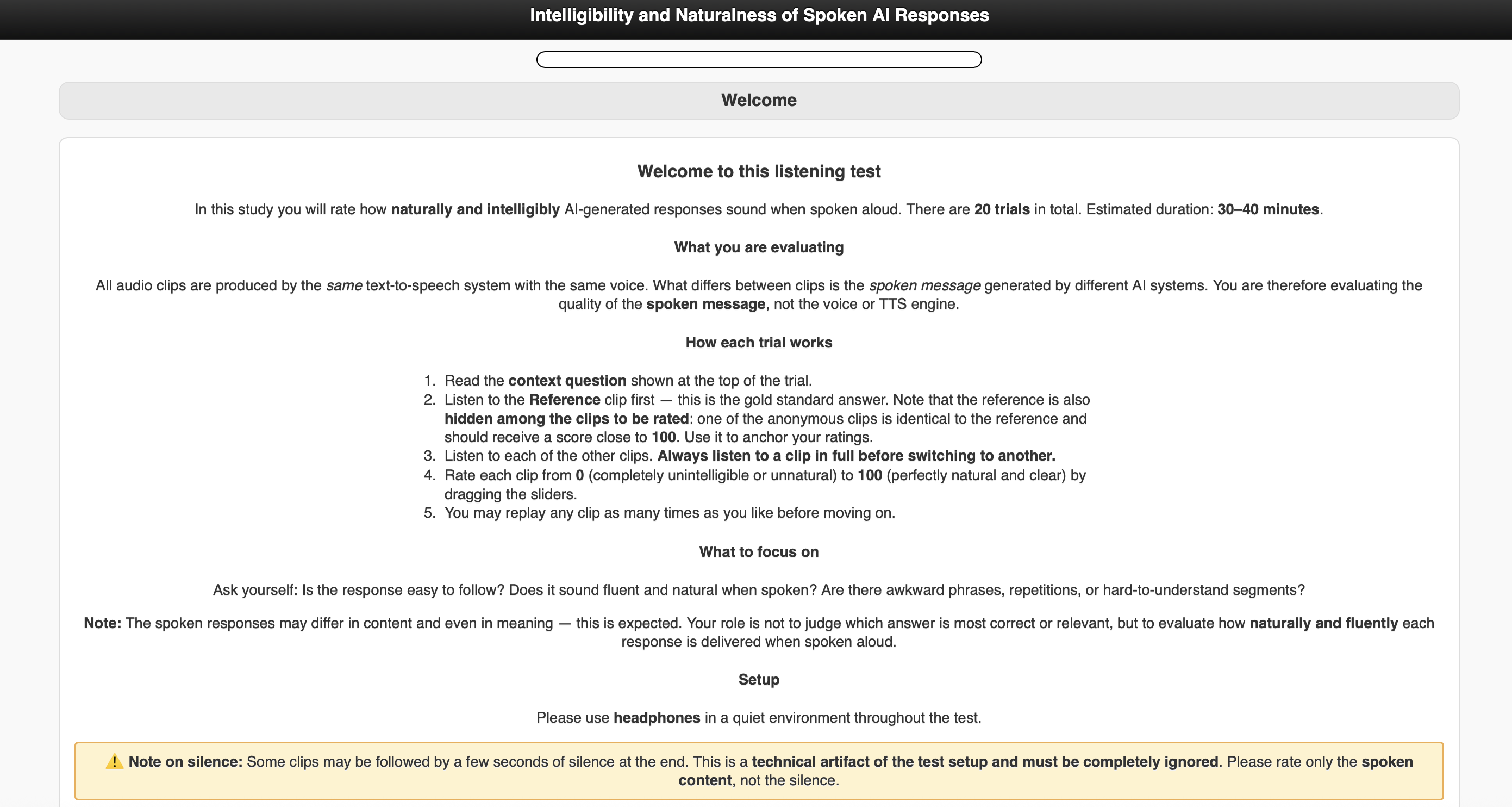}
  \caption{Welcome and instructions page shown to participants at the
  start of the MUSHRA listening test.}
  \label{fig:prolific-welcome}
\end{figure*}

\begin{figure*}[t]
  \centering
  \includegraphics[width=\linewidth]{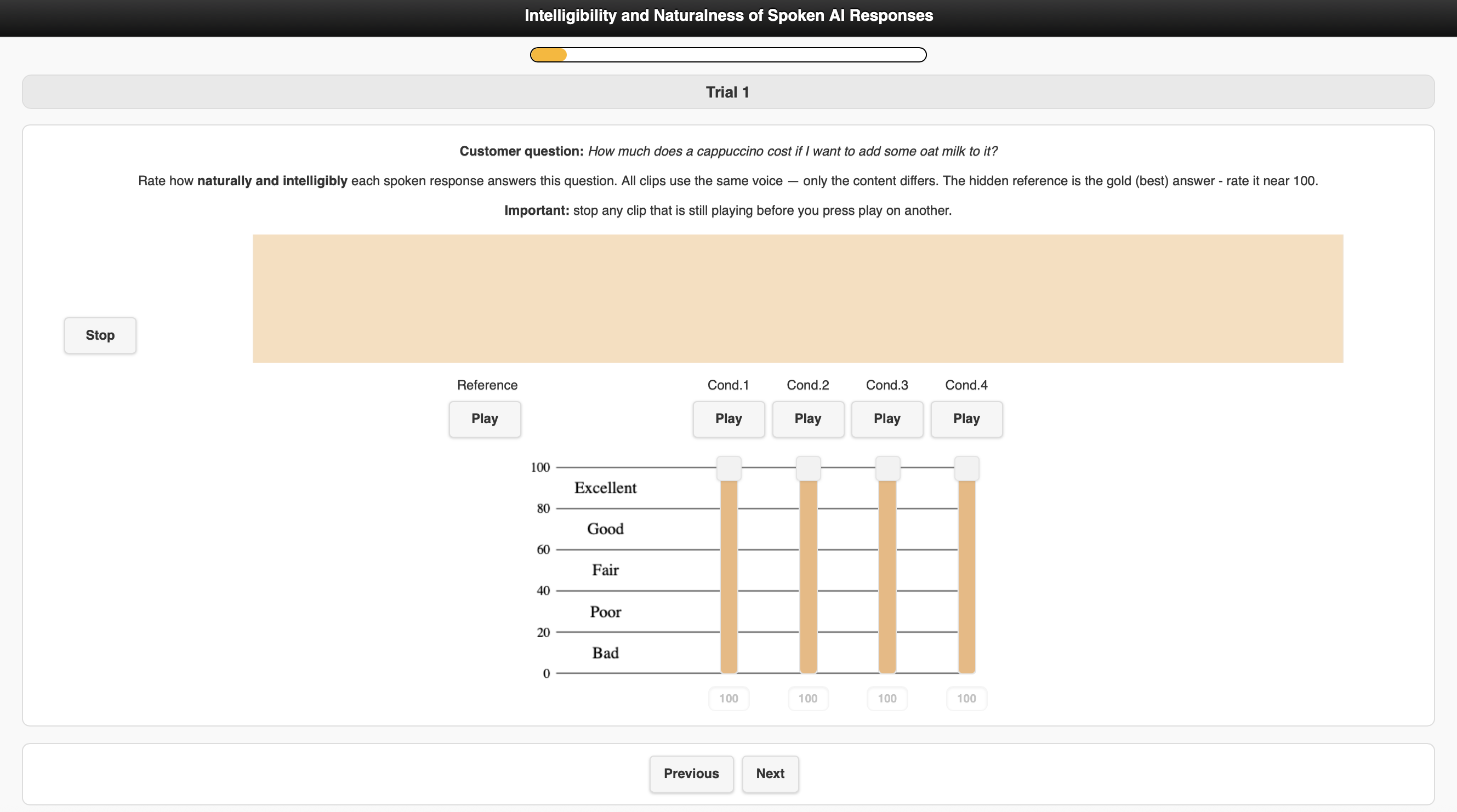}
  \caption{Example MUSHRA trial. The question from CORA is shown at the
  top; the labeled \emph{Reference} (Oracle) clip is on the left; the
  four anonymous, shuffled conditions (Prompting, DPO, FaST, and hidden Oracle) are rated on 0--100 sliders.}
  \label{fig:prolific-trial}
\end{figure*}

\paragraph{Instructions shown to participants.} The exact text rendered
on the welcome page (Figure~\ref{fig:prolific-welcome}) is reproduced
below.

\begin{quote}\small\itshape
Welcome to this listening test. In this study you will rate how
naturally and intelligibly AI-generated responses sound when spoken
aloud. There are 20 trials in total. Estimated duration: 30--40
minutes.\\[2pt]
\textbf{What you are evaluating.} All audio clips are produced by the
same text-to-speech system with the same voice. What differs between
clips is the spoken message generated by different AI systems. You are
therefore evaluating the quality of the spoken message, not the voice
or TTS engine.\\[2pt]
\textbf{How each trial works.} (1) Read the context question shown at
the top of the trial. (2) Listen to the Reference clip first -- this is
the gold-standard answer; the reference is also hidden among the clips
to be rated, and the matching anonymous clip should receive a score
close to 100. (3) Listen to each of the other clips, in full, before
switching. (4) Rate each clip from 0 (completely unintelligible or
unnatural) to 100 (perfectly natural and clear) by dragging the
sliders. (5) You may replay any clip as many times as you like.\\[2pt]
\textbf{What to focus on.} Is the response easy to follow? Does it
sound fluent and natural when spoken? Are there awkward phrases,
repetitions or hard-to-understand segments? The spoken responses may
differ in content and even in meaning -- this is expected. Your role
is not to judge which answer is most correct or relevant, but how
naturally and fluently each response is delivered when spoken
aloud.\\[2pt]
\textbf{Setup.} Please use headphones in a quiet environment throughout
the test.\\[2pt]
\textbf{Note on silence.} Some clips may be followed by a few seconds
of silence at the end. This is a technical artefact of the test setup
and must be completely ignored. Please rate only the spoken content.
\end{quote}

\noindent Each MUSHRA trial (Figure~\ref{fig:prolific-trial}) repeated a
one-sentence reminder: \emph{``Rate how naturally and intelligibly each
spoken response answers this question. All clips use the same voice --
only the content differs. The hidden reference is the gold (best)
answer -- rate it near 100. Stop any clip that is still playing before
you press play on another.''}

\paragraph{Participant pool and reward.} The Prolific study targeted
English-as-first-language participants for a reward of \pounds 12
($\approx$ \pounds 18/hour). Each participant rated all 20 questions $\times$ 4
conditions in a single session (estimated to approximately 40~minutes). 

\paragraph{Consent and data use.} Two consent mechanisms apply. First,
Prolific workers explicitly accept the platform's Participant Terms at
registration, which authorize their submitted responses to be used by
researchers for the studies they take. Second, the welcome page
(Figure~\ref{fig:prolific-welcome}) described the purpose of the study,
the nature of the stimuli, and the requirement to use headphones in a
quiet environment; clicking through the page constituted informed
consent. The data recorded for each participant consist solely of the
per-trial slider ratings, per-trial timestamps, the anonymous Prolific
ID (used only to release payment, not redistributed), and the optional
free-text comment collected on the final page. No personally
identifying information, demographic data, or audio recording from the
participant is collected.

\paragraph{Ethics.} The study was not submitted to an institutional
review board: the task (listening to short synthetic speech clips and
moving 0--100 sliders) involves no sensitive content, no deception, no
biometric data, and no personal information. Participants could
withdraw at any moment without penalty, and all of them, including
those who failed the quality-control checks described in the next
section, were paid above both the UK National Living Wage and
Prolific's recommended minimum. The underlying conversational stimuli
are derived from the synthetic CORA dataset created for this paper.

\paragraph{Quality control.} Three automatic checks were applied to each
submission:
\begin{itemize}
  \item \emph{Hidden-reference check:} the labelled reference clip must
        be rated $\ge 80$ on more than 80\% of the 20 trials (fail otherwise).
  \item \emph{Engagement check:} the three anonymous conditions must not all
        carry identical scores on more than 20\% of trials (fail otherwise).
  \item \emph{Time flag:} mean time per trial must lie in $[5, 300]$ seconds
        (flagged, not excluded).
\end{itemize}

\paragraph{Two-batch collection.} The study was launched in two batches to
eventually reach 14 participants who passed quality control (QC).
\textbf{Batch~1} (2026-05-12) recruited 15 participants; 11 passed QC and 4
failed the hidden-reference check. All 15 were paid as per Prolific's
good-faith policy, but the 4 failing submissions were excluded from
analysis.
\textbf{Batch~2} (2026-05-13) was a top-up of 5 participants, with batch-1 participants excluded via Prolific prescreening. Of the 5 submissions, 1 failed the hidden-reference QC check and 1 was an accidental double-submission (identical scores and session UUID); the remaining 3 passed QC.
The combined valid sample after QC therefore contains $N=14$ participants
(11 from batch~1, 3 from batch~2).

\begin{figure*}[t]
  \centering
  \includegraphics[width=0.91\linewidth]{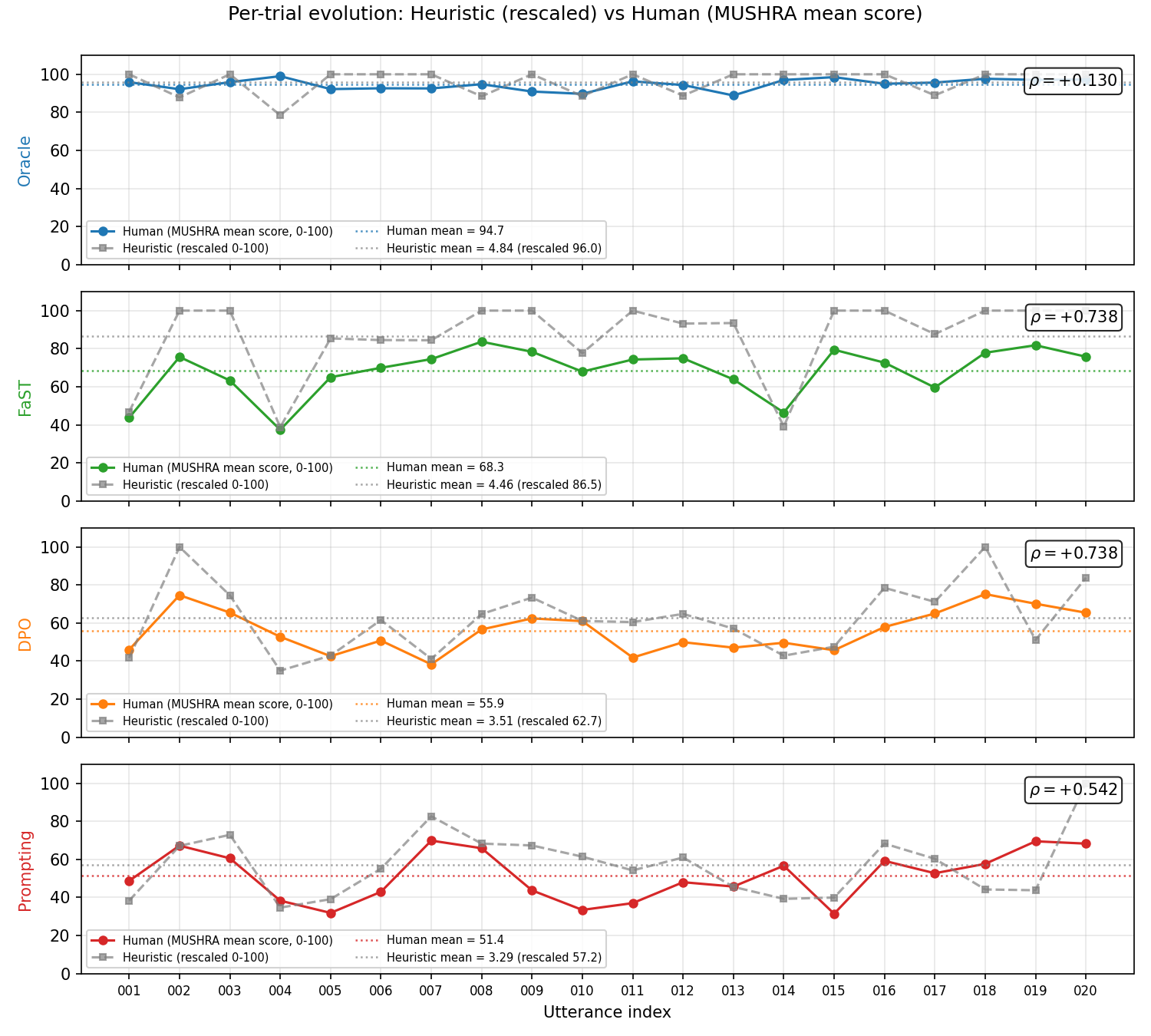}
  \caption{Per-trial Heuristic (rescaled to 0--100) and Human MUSHRA mean score across the 20 utterances of the listening study, one panel per system. The Spearman $\rho$ between the two series is reported in each panel; dotted lines show the per-system means across the 20 trials.}
  \label{fig:mushra_vs_heuristic_per_trial}
\end{figure*}

\paragraph{Heuristic vs Human MUSHRA correlation.} As a final sanity check, we examine the agreement between the cheap and automatic Heuristic TTS-friendliness score (introduced in Section~\ref{sec:data}) and the human ratings collected here. For each of the 20 utterances and each of the 4 systems we compute the mean MUSHRA rating across the $N=14$ approved participants and pair it with the Heuristic score of the corresponding spoken text (Figure~\ref{fig:mushra_vs_heuristic_per_trial}). The pooled Spearman correlation across the $4\times20=80$ (utterance, system) pairs is $\rho=+0.84$ ($p\ll 0.001$); per-system correlations are $\rho_{\text{FaST}}=+0.74$, $\rho_{\text{DPO}}=+0.74$ (both $p<0.001$), $\rho_{\text{Prompting}}=+0.54$ ($p=0.014$), and $\rho_{\text{Oracle}}=+0.13$ (non-significant -- a ceiling effect, since the human-written reference has near-maximal Heuristic and MUSHRA scores on every trial). This confirms the Heuristic metric as a reliable predictor of how naturally 
a spoken message is perceived when rendered by an off-the-shelf TTS system.

\section{Dataset Examples}
\label{app:examples}

Table~\ref{tab:appendix_examples} shows one illustrative example per dataset 
with its associated TTS-friendliness and helpfulness scores.

\begin{table*}[!htbp]
  \centering\footnotesize
  \scalebox{0.94}{
  \begin{tabularx}{1.05\textwidth}{l X l X c c}
    \toprule
    Dataset & Context & Type & Response & TTS-friendliness $\uparrow$ & Helpfulness $\uparrow$ \\
    \midrule
    CORA
      & Before I order, could you confirm if the cardamom maple latte is available throughout the day?
      & chosen
      & Yes, our cardamom maple latte is available all day from six in the morning until eight in the evening. This seasonal special is a popular choice, especially with its delightful cinnamon dusting. Feel free to pair it with one of our freshly baked pastries for a complete treat.
      & 5.00 & 5 \\
      &
      & rejected
      & Yes, the cardamom maple latte is available all day during our store hours from 6:00~AM to 8:00~PM. For more details, visit coffeehouse.com/menu. Pairing it with a pastry is a great idea. Check coffeehouse.com/order for pastries.
      & 2.51 & 5 \\
    \midrule
    Recipe
      & \parbox[t]{\hsize}{%
        How should I cook Neapolitan Pork Chops\\
        Here is the list of ingredients for this recipe:\\
        - 6 (1/2-inch thick) pork chops\\
        - 1/2 tsp. pepper\\
        - 1/2 tsp. oregano\\
        - 1 (14 oz.) can Italian tomatoes\\
        - 1 Tbsp. olive oil\\
        - 1/2 tsp. garlic powder\\
        - 1/4 tsp. sugar\\
        - 1/2 c. sliced olives
      }
      & chosen
      & Sprinkle six pork chops, which are each one half inch thick, with one half teaspoon of pepper and one half teaspoon of oregano, then brown them in one tablespoon of olive oil. Next, add one can of Italian tomatoes, which is fourteen ounces, and let the mixture simmer for thirty minutes, or until the chops are tender. Finally, stir in one half cup of sliced olives and serve the dish over noodles.
      & 5.00 & 4 \\
      &
      & rejected
      & Sprinkle 6 (1/2-inch thick) pork chops with 1/2 tsp. pepper, 1/2 tsp. oregano, and 1/2 tsp. garlic powder; brown in 1 Tbsp. olive oil. Add 1 (14 oz.) can Italian tomatoes; simmer for 30 minutes, or until chops are tender. Stir in 1/2 c. sliced olives and serve over noodles.
      & 2.29 & 4 \\
    \bottomrule
  \end{tabularx}
  }
  \caption{One illustrative $\langle$\textit{context} question, \textit{chosen} response, \textit{rejected} response$\rangle$ tuple from each dataset. We also report the TTS-friendliness and Helpfulness score for each response.}
  \label{tab:appendix_examples}
\end{table*}


\section{Details on the Heuristic TTS-Friendliness Metric}
\label{app:heuristic-metric}

Given a candidate text $t$, the proposed heuristic metric script counts several surface patterns known
to be problematic for TTS, aggregates their length-normalised frequencies
into a risk value, and maps it to a 1--5 score (higher meaning more TTS-friendly).
Each class $k$ is matched by a Python regex with a fixed weight $w_k$
(Table~\ref{tab:heuristic-patterns}).

\begin{table*}[t]
\centering
\small
\begin{tabular}{lll}
\toprule
\textbf{Class $k$} & \textbf{Regex} & $w_k$ \\
\midrule
\texttt{acronym\_caps}        & \verb![A-Z]{2,}!                                            & 0.5 \\
\texttt{numbers}              & \verb!\b\d+(?:[\.,]\d+)?\b!                                 & 0.5 \\
\texttt{url\_like}            & \verb!(https?://|www\.|[A-Za-z0-9]+\.[A-Za-z]{2,})(/\S*)?! & 1.0 \\
\texttt{mix\_alphanum\_punct} & \verb![A-Za-z0-9]+[-_/:.][A-Za-z0-9]+!                      & 0.5 \\
\texttt{symbols}              & \verb![\^\*\+\=\~\|\@\#\$%&]!                               & 0.5 \\
\texttt{equation}             & \verb![A-Za-z]\s*=\s*[^\s]+!                                & 1.0 \\
\bottomrule
\end{tabular}
\caption{Regexes and weights for the Heuristic TTS-Friendliness metric. URL- and equation-like patterns are weighted
twice as much, being the most disruptive for naive TTS front-ends. Emails,
code identifiers, file paths, dates, prices and units are captured
indirectly by combinations of \texttt{url\_like},
\texttt{mix\_alphanum\_punct}, \texttt{numbers}, and \texttt{symbols}.}
\label{tab:heuristic-patterns}
\end{table*}

\paragraph{Score.} For each class $k$, let $n_k$ be the number of
non-overlapping matches of regex $k$ in $t$ and $m_k$ the total number
of characters they cover (Python \texttt{re}, case-sensitive defaults;
classes are scored independently, so overlaps between classes are
allowed). Let $L=\max(|t|,20)$, the floor of $20$ preventing very short
texts from being over-penalized. The per-class score, the aggregate
risk, and the final 1--5 score are defined as follows:
\begin{align}
s_k &= w_k \left( n_k + 0.25\,\tfrac{m_k}{L} \right), \\
\mathrm{R} &= \frac{\sum_{k} s_k}{1 + L/200}, \\
\mathrm{S}_\textsc{Heur} &= 5 - 4\,\tfrac{\mathrm{R}}{\mathrm{R}+2}\;\in\;[1,5].
\end{align}
The count $n_k$ rewards \emph{how often} a risky pattern appears, the
fraction $m_k/L$ \emph{how much} of the text it covers. The $1+L/200$
denominator limits accumulation in long answers. The final map
saturates at $5$ when no risky pattern is present ($\mathrm{R}=0$) and
approaches $1$ as risk grows, so higher scores mean TTS-friendlier text.

\paragraph{Calibration.} The weights, the $0.25$ coverage coefficient, the
floor $L\!\geq\!20$ and the $1+L/200$ normalizer were set by inspection on
a small held-out sample so that plain natural-language answers score close
to 5 and a single URL or equation drops the score by at least one full
point. No evaluation set from the paper was used for tuning. The exact
script to compute the heuristic score will be released 
for reproducibility.

\section{Hyperparameters and Baseline Details}
\label{app:hyperparams}

The hyperparameters used in our experiments are detailed in Table~\ref{tab:hyperparams}. They are directly based on the hyperparameters used in~\citet{fast}, which we found to yield satisfactory results in our pilot experiments on CORA and Recipe.

For the PolyNorm baseline~\citep{Wong2025tn} discussed in Section~\ref{sec:text-normalization}, given the absence of public code, we re-implemented the approach using the prompt provided in the paper and used English examples from PolyNorm-Bench\footnote{\url{https://github.com/apple/ml-speech-polynorm-bench}} as few-shot examples. To roughly match the number of few-shot examples specified in the paper, we selected 4 examples for each of the 27 categories, resulting in a 108-shot setting.

\begin{table*}[t]
\centering
\small
\scalebox{0.95}{
\begin{tabular}{lp{13.5cm}}
\toprule
\textbf{Approach} & \textbf{Hyperparameters} \\ \midrule
\multicolumn{2}{l}{\textit{Reward model training}} \\
\arrayrulecolor{lightgray}\midrule
RM & {learning\_rate = 1.41e-5}, {batch\_size = 16}, {num\_train\_epochs = 2} \\
FaRM & {learning\_rate = 0.1}, {max\_iter = 500}, {tolerance = 0.1} \\ \midrule
\multicolumn{2}{l}{\textit{Generation model fine-tuning}} \\ \midrule
SFT & {learning\_rate = 1.41e-5}, {batch\_size = 16}, {num\_train\_epochs = 10} \\
DPO & {learning\_rate = 5.0e-6}, {batch\_size = 16}, {num\_train\_epochs = 10}, {beta = 0.1} \\
GRPO & {num\_samples = 10}, {train\_temperature = 1.2}, {train\_top\_p = 0.9}, {learning\_rate = 1.41e-5}, {batch\_size = 16}, {num\_train\_iters = 5}, {num\_train\_epochs = 5}, {beta = 0.01} \\
RFT & {num\_samples = 10}, {train\_temperature = 1.2}, {train\_top\_p = 0.9}, {learning\_rate = 1.41e-5}, {batch\_size = 16}, {num\_train\_iters = 5}, {num\_train\_epochs\_per\_iter = 5} \\ \midrule
\multicolumn{2}{l}{\textit{Evaluation}} \\ \midrule
Sampling & {eval\_temperature = 0.7}, {eval\_top\_k = 20}, {eval\_top\_p = 0.8}, {max\_length = 1024} \\
    
\arrayrulecolor{black}\bottomrule
\end{tabular}
}
\caption{Hyperparameters adopted for reward model training, generation model fine-tuning, and evaluation.}
\label{tab:hyperparams}
\end{table*}

\section{Computational Infrastructure}

The fine-tuning of the generation models and the traditional reward model (RM) was conducted on a single A100 GPU. The training on the five data splits took between 1 hour and 10 hours for CORA, and between 2 hours and 20 hours for Recipe~-- these numbers depending both on the approach and the data regime (10 vs. 100 training samples). The weight learning of the feature-aware reward model (FaRM) was done on CPU only, given the very small number of parameters to learn (equal to the number of features $F=40$). 

\section{Additional Results}

\subsection{Generation Results with SmolLM3-3B}
\label{app:results-smollm}

\begin{figure*}[t]
    \centering
    \begin{subfigure}[b]{0.49\textwidth}
        \includegraphics[width=\textwidth]{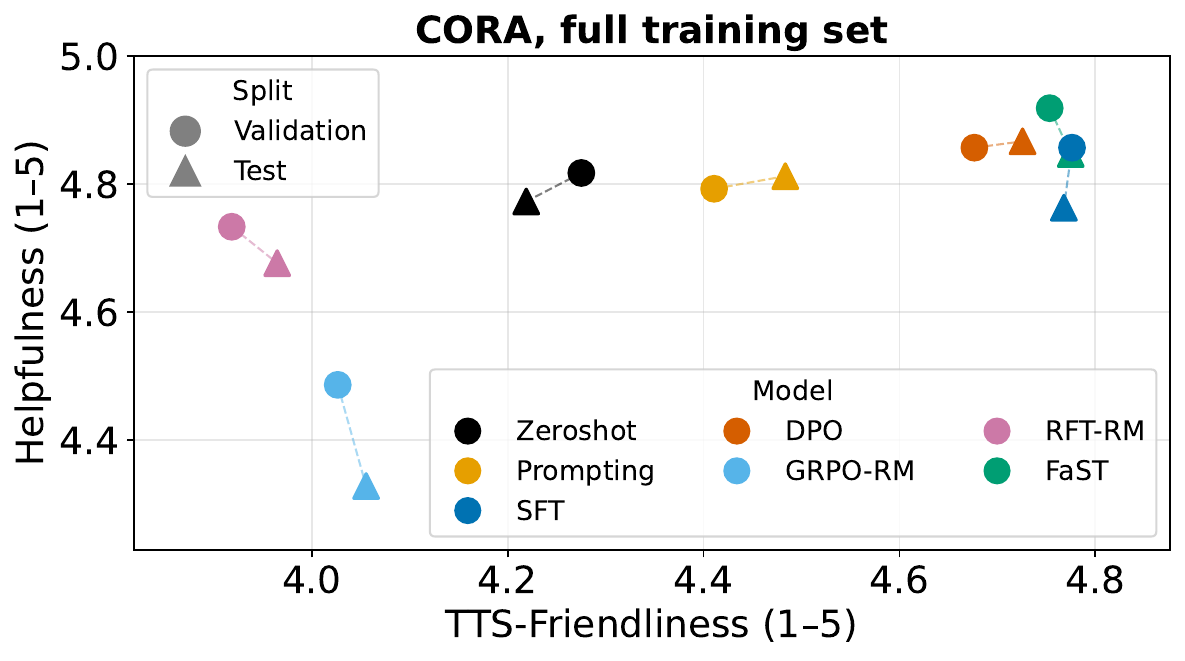}
    \end{subfigure}
    \hfill
    \begin{subfigure}[b]{0.49\textwidth}
        \includegraphics[width=\textwidth]{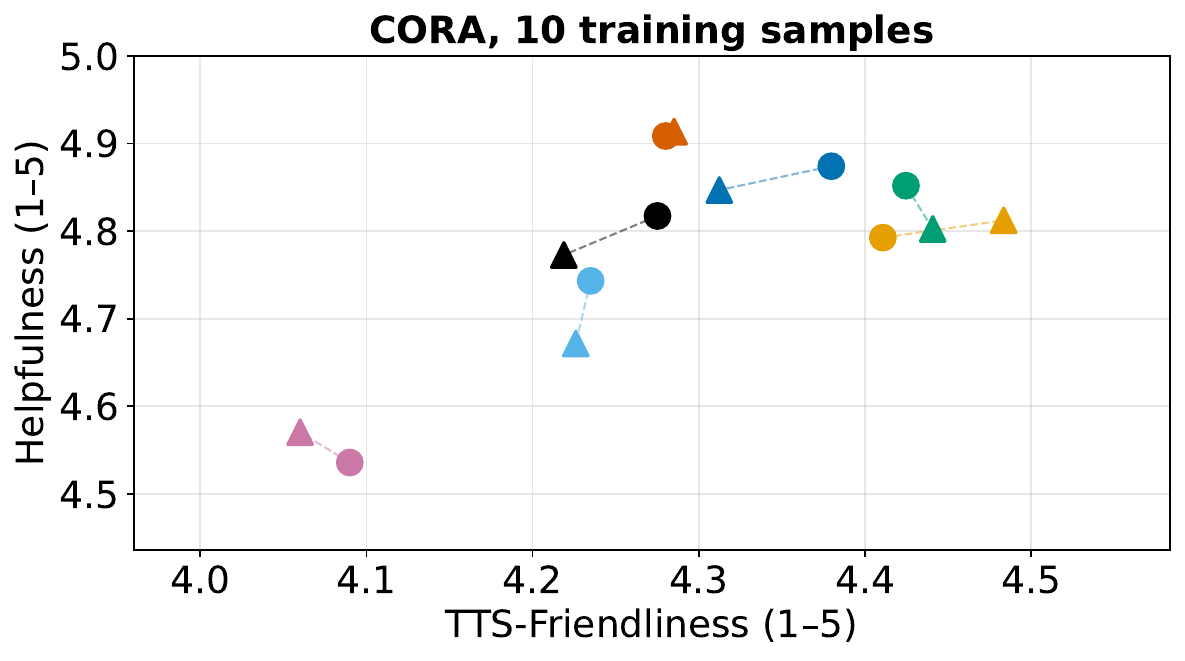}
    \end{subfigure}
    \vspace{0.1cm}
    \begin{subfigure}[b]{0.49\textwidth}
        \includegraphics[width=\textwidth]{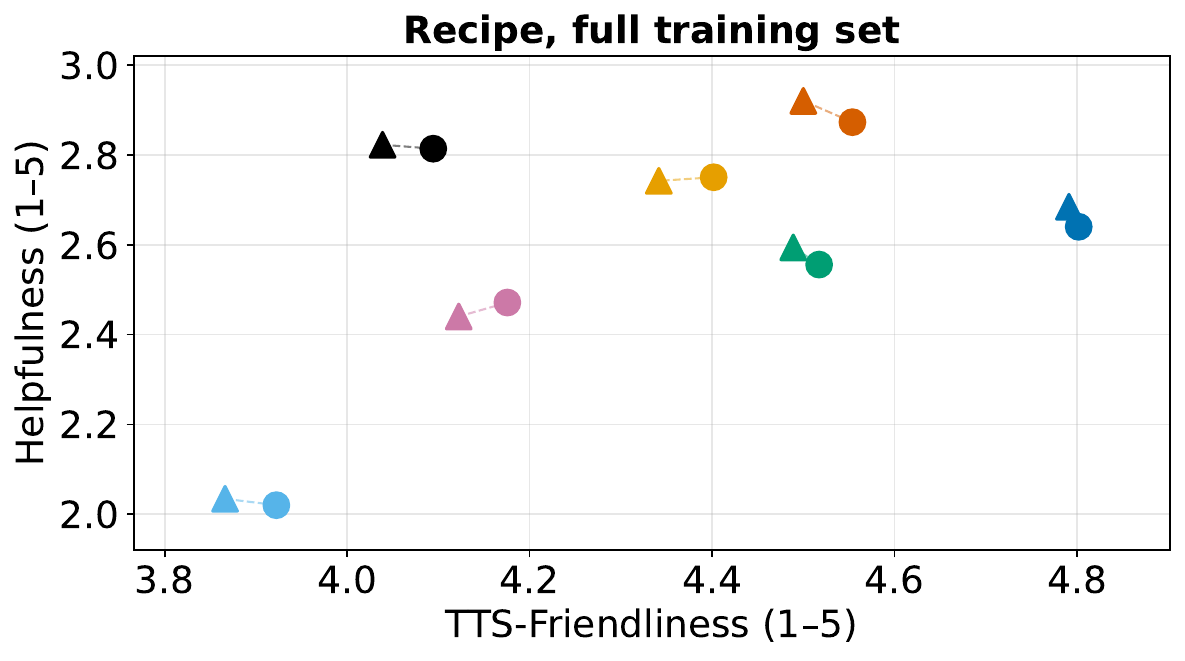}
    \end{subfigure}
    \hfill
    \begin{subfigure}[b]{0.49\textwidth}
        \includegraphics[width=\textwidth]{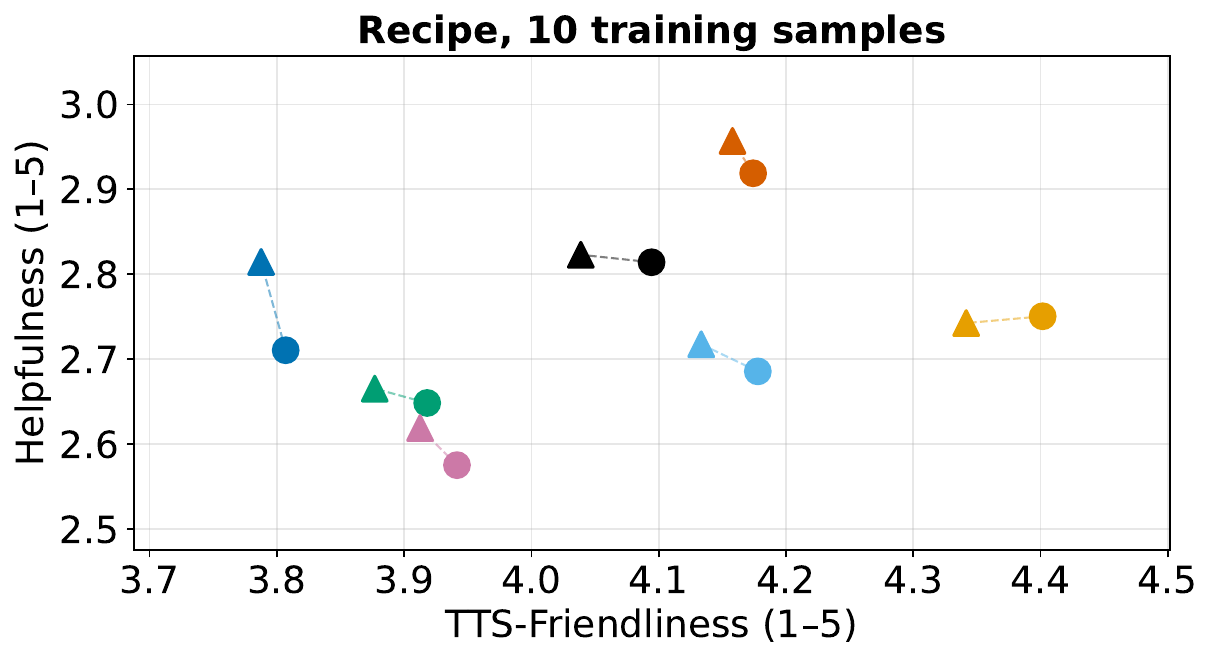}
    \end{subfigure}
    \caption{Comparison of generation approaches on the tradeoff between TTS-Friendliness and Helpfulness (top-right is better). The base model used is SmolLM3-3B. To improve readability, the x-axis and y-axis have been adjusted for each plot based on visible data points; they are not identical across the full-set and 10-sample settings.}
    \label{fig:results-smollm}
\end{figure*}

The results obtained with SmolLM3-3B\footnote{\url{https://huggingface.co/HuggingFaceTB/SmolLM3-3B}} (Figure~\ref{fig:results-smollm}) are broadly aligned with the trends observed with Qwen3-4B in Figure~\ref{fig:results-qwen}, suggesting that the competitiveness of FaST is not tied to a single model family.\footnote{In this experiment, the reward models FaRM and RM used respectively in FaST and GRPO-RM/RFT-RM are the same as those used in the experiments with Qwen3-4B. Only the generation model to be fine-tuned was changed to SmolLM3-3B, to limit compounding variations that would make interpretability of the results more challenging.} Across both CORA and Recipe, FaST again provides one of the strongest overall tradeoffs between TTS-friendliness and Helpfulness. 

On CORA, FaST remains consistently near the Pareto frontier in both the 10-sample and full-data settings, achieving among the highest TTS-friendliness scores while preserving strong helpfulness. The Recipe dataset exhibits a slightly different pattern. While FaST remains highly competitive in the full-training setting, we note a degradation when 10 training samples are used. We hypothesize that this could be due to the fact that the random set of 10 training samples might contain examples with a weak training signal (i.e., a low contrast between the TTS-friendly and TTS-unfriendly responses). Nonetheless, we observe that in this case FaST was still able to improve its TTS-friendliness score over SFT, suggesting that the features captured by FaST overall helped produce samples more appropriate for spoken delivery.

\subsection{Comparison of Generation Lengths}
\label{app:length} 

Table~\ref{tab:lengths} compares the length of the responses generated by the different approaches on CORA and Recipe. The length is measured by the number of characters in the generated response, which we average over all the validation and test responses. As pointed out in our \textit{Limitations} section, we can observe that FaST tends to output responses that are overall lengthier than most approaches. This may be explained by the fact that FaST's discovered features and learned weights associate conciseness and brevity as negatively correlated with TTS-friendliness (see Tables~\ref{tab:cora-features40} and~\ref{tab:recipe-features40}). This, in turn, is likely caused by the fact that TTS-unfriendly responses are overall shorter due to their more compact formatting and high abbreviation usage.

\begin{table}[t]
    \centering
    \small
    \begin{tabular}{lrr}
    \toprule
    \multirow{2}{*}{\textbf{Approach}} & \textbf{CORA} & \textbf{Recipe} \\
    & \textbf{Gen. length} & \textbf{Gen. length} \\
    \midrule
    \multicolumn{3}{c}{\textit{Training-free}} \\
    \midrule
    Zeroshot      & 185.6 & 480.9 \\
    Prompting        & 178.9 & 547.6 \\
    Oracle & 231.0 & 813.0 \\
    \midrule
    \multicolumn{3}{c}{\textit{10 training samples}} \\ 
    \midrule
    SFT           & 205.2 & 571.4 \\
    DPO           & 242.9 & 552.3 \\
    GRPO-RM       & 183.4 & 512.4 \\
    RFT-RM        & 172.0 & 568.1 \\
    FaST     & 254.7 & 1333.4 \\
    \midrule
    \multicolumn{3}{c}{\textit{100 training samples}} \\
    \midrule
    SFT           & 221.0 & 681.6 \\
    DPO           & 292.3 & 733.4 \\
    GRPO-RM       & 1339.2 & 1028.4 \\
    RFT-RM        & 205.6 & 694.0 \\
    FaST      & 434.5 & 901.9 \\
    \bottomrule
    \end{tabular}
    \caption{Average generation lengths on the validation and test sets combined. The length is measured as the number of characters in the generated response.}
    \label{tab:lengths}
\end{table}

\subsection{Verbosity Control in FaST}
\label{app:verbosity-control} 

To address FaST's length issue discussed in our \textit{Limitations} section and in App.~\ref{app:length}, we investigate the suggested trick which consists in manually editing the weight of the feature controlling verbosity. This feature corresponds to \textit{conciseness} on CORA and \textit{brevity} on Recipe (see Tables~\ref{tab:cora-features40} and~\ref{tab:recipe-features40}), which both carry negative weights: $-0.12$ and $-0.25$, respectively. We thus  compare FaST in its default version (with the full training set and using Qwen3-4B as the base model) against the variant where we zero out the conciseness feature, on the CORA dataset. The results are reported in Table~\ref{tab:verbosity-control}. These results confirm that our suggested fix works as intended, notably reducing the generation length while preserving the TTS-friendliness heuristic score. We also note that it might be possible to further reduce the generation length by using a small positive weight for the conciseness feature, instead of setting it to zero~-- although this might lead to penalizing TTS-friendliness if the magnitude of the positive weight is chosen too large.

\begin{table}[t]
\centering
\scalebox{0.9}{
\begin{tabular}{lrr}
\toprule
\textbf{Approach} & \textbf{Gen. length} & \textbf{TTS-friendliness} \\
\midrule
FaST (default) & 434.6 & 4.73 \\
FaST ($\lambda_{\textit{conciseness}}=0$) & 335.6 & 4.74 \\
\bottomrule
\end{tabular}
}
\caption{Impact of zeroing out the conciseness feature weight in FaST, in terms of generation length and heuristic TTS-friendliness score.}
\label{tab:verbosity-control}
\end{table}

\subsection{Generated Samples}
\label{app:generations} 

We provide in Table~\ref{tab:recipe-samples-all} the samples generated by the different approaches on one context for the Recipe dataset.













\begin{table*}[t]
\centering
\small
\scalebox{0.89}{
\begin{tabular}{@{}p{1.7cm}p{13.3cm}@{}}
\toprule
\multicolumn{2}{@{}c@{}}{\textbf{Recipe}} \\
\midrule

\multicolumn{2}{@{}p{15cm}@{}}{%
\parbox[t]{15cm}{%
\textbf{Context:} How should I cook Neapolitan Pork Chops\\
Here is the list of ingredients for this recipe:\\
- 6 (1/2-inch thick) pork chops\\
- 1/2 tsp. pepper\\
- 1/2 tsp. oregano\\
- 1 (14 oz.) can Italian tomatoes\\
- 1 Tbsp. olive oil\\
- 1/2 tsp. garlic powder\\
- 1/4 tsp. sugar\\
- 1/2 c. sliced olives
}%
} \\
\vspace{0.2cm}

\textbf{Responses:} & \\

\textit{Zeroshot:} &
Heat olive oil in a large skillet over medium-high heat. Add pork chops and sear until golden brown on both sides, about 3--4 minutes per side. Remove chops and set aside. In the same skillet, add garlic powder, oregano, pepper, and sugar; stir to coat. Add sliced olives and tomatoes, then bring to a simmer. Return pork chops to the skillet, cover, and cook for 10--12 minutes, or until pork is tender and internal temperature reaches 145°F. Serve warm. \\

\textit{Prompting:} &
First, preheat your oven to 375 degrees Fahrenheit.
Place the pork chops on a baking sheet.
Add one tablespoon of olive oil to the chops.
Season the pork chops with one-half teaspoon of pepper, one-half teaspoon of oregano, one-half teaspoon of garlic powder, and one-quarter teaspoon of sugar.
Add the sliced olives to the baking sheet.
Pour the Italian tomatoes over the pork chops and olives.
Bake the pork chops for 25 minutes.
Check the chops to make sure they are cooked through.
Serve warm. \\

\textit{SFT:} &
Preheat your oven to three hundred fifty degrees. In a shallow dish, place six one half inch thick pork chops, then season them with one half teaspoon of pepper, one half teaspoon of oregano, and one half teaspoon of garlic powder. In a three quart casserole dish, combine one can, or fourteen ounces, of Italian tomatoes, one tablespoon of olive oil, one quarter teaspoon of sugar, and one half cup of sliced olives, stirring well. Place the pork chops in the tomato mixture and bake for one hour and fifteen minutes, or until the pork is tender and the sauce is thick. \\

\textit{DPO:} &
Preheat your oven to 375°F. Season the pork chops generously with pepper, oregano, garlic powder, and a pinch of sugar. Heat one tablespoon of olive oil in a large skillet over medium heat, then sear the pork chops until golden brown on both sides, about 3 to 4 minutes per side. Remove the chops from the skillet and set them aside. In the same skillet, add the sliced olives and cook them for 1 to 2 minutes until they soften slightly. Pour in the canned Italian tomatoes and stir to combine, then bring the mixture to a gentle simmer. Return the pork chops to the skillet, nestling them into the sauce, and bake for 15 to 20 minutes, or until the pork is cooked through and reaches an internal temperature of 145°F. Serve warm, with the sauce clinging to the chops. \\

\textit{GRPO-RM:} &
Warm oven to 400°F. Heat olive oil in a large skillet over medium heat. Add pork chops and season generously with pepper, oregano, and garlic powder; cook 3-4 minutes per side until golden and slightly crisp. In the meantime, crush Italian tomatoes with a fork, and add to the skillet along with sliced olives, sugar, and 1/4 cup of water (or broth). Cook 10 minutes, stirring occasionally, until tomatoes break down into a thick sauce. Add the pork chops to the sauce, and cook 5 minutes more, until the sauce is heated through and the chops are warmed. Remove chops from skillet and let rest 5 minutes before serving. Avoid overcooking, as pork chops can dry out easily. The sauce will thicken as it cools, so serve slightly warm or at room temperature. Use a meat thermometer to ensure the internal temperature reaches 145°F, then rest before serving. This method ensures juicy, flavorful pork chops with a rich, tomato-based sauce. \\

\textit{RFT-RM:} &
Place six one-half-inch thick pork chops in a single layer in a greased 8x8 inch casserole dish. In a small bowl, mix one half teaspoon of pepper, one half teaspoon of oregano, one half teaspoon of garlic powder, one quarter teaspoon of sugar, and one cup of sliced olives. Sprinkle two-thirds of this mixture over the pork chops, season the tops of the chops with the remaining seasoning, and dot with one tablespoon of olive oil. Cover and bake at four hundred degrees Fahrenheit for thirty minutes. In a separate bowl, combine one can of Italian tomatoes and the remaining seasoning, pour this over the pork chops, and continue baking for an additional fifteen minutes. \\

\textit{FaST:} &
Pat six pork chops dry, then season both sides with salt, pepper, oregano, and garlic powder. In a large skillet, heat olive oil over medium-high heat until shimmering; add pork chops and sear until golden and crusty on both sides, about 3 minutes per side. Remove chops and add sliced olives to the same skillet; sauté until fragrant, about 1 minute. In the same skillet, stir together Italian tomatoes, sugar, and a splash of water; bring to a gentle boil, then reduce heat and simmer uncovered for 15 minutes. Return pork chops to the skillet, baste with the sauce, and simmer for another 5 minutes. Rest chops for 5 minutes before serving, drizzled with reserved sauce. \\

\bottomrule
\end{tabular}
}
\caption{Example generations from multiple approaches for the Recipe dataset.}
\label{tab:recipe-samples-all}
\end{table*}

\section{Discovered Features}
\label{app:features}

Tables~\ref{tab:cora-features40} and~\ref{tab:recipe-features40} report the full sets of features discovered by FaST on CORA and Recipe respectively, together with their learned weights. The description associated to each feature in the table was discovered automatically during the feature discovery step of FaST (see Section~\ref{sec:fast} for more details). Across both datasets, the sign and magnitude of the weights align closely with the linguistic properties known to affect TTS performance~\citep{Zhang2024tn,Wong2025tn}. Features capturing compact written conventions~-- such as \textit{use\_of\_numeric\_formatting} ($-0.50$ on CORA) and \textit{numeral\_symbol\_usage} ($-0.45$ on Recipe)~-- receive strongly negative weights, which is consistent with the sensitivity of TTS systems to symbols, digits, and abbreviations that lack a canonical spoken form. Conversely, features associated with natural spoken delivery~-- \textit{natural\_conversational\_tone} ($+0.33$) and \textit{narrative\_prose\_style} ($+0.76$ on Recipe)~-- are positively rewarded, reflecting the expectation that fluent, prose-like text is more likely to be associated with TTS-friendliness in our preference datasets.

The features discovered on the two datasets also highlight different patterns that reveal domain-specific phenomena. On CORA, \textit{external\_referral} ($-0.24$) and \textit{technical\_code\_reference} ($-0.18$) carry meaningful negative weights, capturing the tendency of written customer-service responses to include URLs, email addresses, and identifiers that are challenging for TTS systems. On Recipe, \textit{abbreviation\_density} ($-0.58$) and \textit{parenthetical\_annotation} ($-0.35$) dominate, reflecting the heavily compressed shorthand notation typical of recipe writing (e.g., 1 Tbsp., 1/2 tsp.). The fact that FaST recovers these domain-specific patterns automatically~-- without any domain-specific engineering~-- further supports the use of feature-based reward models for TTS-friendly generation across heterogeneous domains.

\begin{table*}[t]
    \centering
    \scalebox{0.78}{
    \begin{tabular}{lrp{12.7cm}}
    \toprule
    \textbf{Feature} & \textbf{Weight} & \textbf{Description} \\\midrule
    use\_of\_numeric\_formatting & \colorvaluegradient{-0.50} & How much does the response rely on digits, symbols, and compact numeric notation such as \$5.25, 6:00 AM, or +\$0.50? \\
    natural\_conversational\_tone & \colorvaluegradient{0.33} & How natural and conversational does the response sound in customer-facing dialogue? \\
    use\_of\_spelled\_out\_numbers & \colorvaluegradient{0.33} & How much does the response prefer fully written-out numbers and amounts in prose? \\
    abbreviation\_density & \colorvaluegradient{-0.32} & How much does the response use abbreviations, shorthand, or compressed forms such as w/, FYI, ASAP, etc., or St.? \\
    external\_referral & \colorvaluegradient{-0.24} & How much does the response redirect the user to external resources such as websites, apps, email, or phone instead of answering fully in place? \\
    friendliness & \colorvaluegradient{0.23} & How friendly and welcoming is the response? \\
    descriptive\_sensory\_language & \colorvaluegradient{0.19} & How much does the response use sensory or evocative language about taste, texture, aroma, or experience? \\
    professional\_polish & \colorvaluegradient{0.18} & How polished and professional is the wording of the response? \\
    technical\_code\_reference & \colorvaluegradient{-0.18} & How much does the response include internal-looking identifiers such as SKU, item IDs, or order codes? \\
    enthusiasm & \colorvaluegradient{0.17} & How much positive energy or enthusiasm does the response convey? \\
    follow\_up\_engagement & \colorvaluegradient{0.16} & How much does the response invite the user to continue the interaction with a follow-up question or offer of help? \\
    policy\_grounding & \colorvaluegradient{-0.15} & How much does the response reflect operational rules, constraints, or store policy language? \\
    conciseness & \colorvaluegradient{-0.12} & How concise is the response while still remaining understandable? \\
    detail\_richness & \colorvaluegradient{0.11} & How much descriptive or supporting detail does the response provide? \\
    persuasive\_recommendation\_style & \colorvaluegradient{0.11} & How strongly does the response try to encourage interest, purchase, or further engagement? \\
    accuracy\_signal\_for\_menu\_facts & \colorvaluegradient{-0.11} & How much does the response present concrete menu facts in a way that appears precise and dependable? \\
    customer\_service\_orientation & \colorvaluegradient{0.11} & How strongly does the response reflect a service-oriented mindset focused on helping the customer smoothly? \\
    fee\_transparency & \colorvaluegradient{0.10} & How clearly does the response explain extra charges, add-on costs, or total pricing implications? \\
    time\_specificity & \colorvaluegradient{-0.10} & How explicitly does the response provide concrete times, hours, or timing constraints? \\
    promotional\_language & \colorvaluegradient{0.10} & How much does the response use promotional or marketing-style phrasing about products or the store experience? \\
    recommendation\_strength & \colorvaluegradient{0.09} & How strongly does the response make a recommendation rather than simply listing options? \\
    self\_containment & \colorvaluegradient{0.09} & How self-contained is the response, meaning the user can act on it without needing another source? \\
    clarity\_of\_wording & \colorvaluegradient{0.09} & How easy is the response to understand on first reading? \\
    response\_formality & \colorvaluegradient{0.09} & How formal is the style of the response? \\
    urgency\_emphasis & \colorvaluegradient{-0.08} & How much does the response stress acting quickly or time sensitivity? \\
    redundancy\_level & \colorvaluegradient{-0.08} & How much does the response repeat information unnecessarily within the same answer? \\
    procedural\_clarity & \colorvaluegradient{0.07} & How clearly does the response explain a process, policy, or sequence of actions? \\
    contextual\_helpfulness & \colorvaluegradient{0.07} & How well does the response tailor its content to the situation implied by the user's context, such as first-time visit, warm day, or dietary need? \\
    menu\_item\_specificity & \colorvaluegradient{0.05} & How specifically does the response describe the item, including ingredients, preparation, or characteristics? \\
    price\_specificity & \colorvaluegradient{-0.05} & How explicitly does the response provide exact pricing information when relevant? \\
    customization\_support & \colorvaluegradient{0.05} & How much does the response support modifications, substitutions, or personalized options? \\
    location\_specificity & \colorvaluegradient{-0.04} & How specifically does the response describe a place, address, or directions? \\
    comparative\_guidance & \colorvaluegradient{0.04} & How much does the response help compare alternatives based on preferences or use cases? \\
    appropriateness\_of\_extra\_information & \colorvaluegradient{0.04} & How appropriate and useful are any extra details beyond the core answer? \\
    actionability & \colorvaluegradient{-0.03} & How actionable is the response in telling the user what to do next? \\
    grammatical\_cleanliness & \colorvaluegradient{0.02} & How grammatically clean and well-formed is the response? \\
    hedging\_vs\_certainty & \colorvaluegradient{0.02} & How confidently does the response present its information rather than hedging or sounding tentative? \\
    completeness\_of\_response & \colorvaluegradient{0.01} & To what extent does the response cover all parts of the user's request? \\
    direct\_answer\_relevance & \colorvaluegradient{0.00} & How directly does the response answer the user's specific question or request without drifting to unrelated details? \\
    dietary\_accommodation\_clarity & \colorvaluegradient{0.00} & How clearly does the response address dietary needs such as vegan, dairy-free, or gluten-free requirements? \\
    \bottomrule
    \end{tabular}
    }
    \caption{Features discovered by FaST on CORA and their learned weights. Features are ordered by the magnitude of their corresponding weight in absolute value.}
    \label{tab:cora-features40}
\end{table*}

\begin{table*}[t]
    \centering
    \scalebox{0.78}{
    \begin{tabular}{lrp{12.7cm}}
    \toprule
    \textbf{Feature} & \textbf{Weight} & \textbf{Description} \\\midrule
    narrative\_prose\_style & \colorvaluegradient{0.76} & To what extent is the choice written in full natural-language prose rather than recipe shorthand? \\
    abbreviation\_density & \colorvaluegradient{-0.58} & How heavily does the choice rely on abbreviations, symbols, and shorthand formatting? \\
    numeral\_symbol\_usage & \colorvaluegradient{-0.45} & To what extent does the choice present quantities with numerals and symbols rather than spelled-out words? \\
    parenthetical\_annotation & \colorvaluegradient{-0.35} & How much does the choice use parenthetical clarifications or side notes? \\
    brevity & \colorvaluegradient{-0.25} & How concise is the wording of the choice? \\
    measurement\_unit\_variety & \colorvaluegradient{0.19} & How varied and explicit are the measurement units used in the choice? \\
    editorial\_commentary & \colorvaluegradient{0.18} & How much does the choice include comments, tips, opinions, or asides beyond core instructions? \\
    redundancy\_level & \colorvaluegradient{0.14} & How much does the choice repeat information that could have been stated more compactly? \\
    readability\_simplicity & \colorvaluegradient{-0.13} & How easy is the choice to read quickly and parse at a glance? \\
    technique\_explanation & \colorvaluegradient{0.12} & How much does the choice explain cooking techniques or methods beyond simply naming them? \\
    formal\_recipe\_register & \colorvaluegradient{-0.12} & How strongly does the choice follow a conventional formal recipe register? \\
    equipment\_guidance & \colorvaluegradient{0.10} & How much does the choice specify cookware, appliances, or tools to use? \\
    formatting\_compactness & \colorvaluegradient{-0.09} & How compressed is the formatting of the choice overall? \\
    serving\_guidance & \colorvaluegradient{0.09} & How much does the choice include serving, plating, or accompaniment suggestions? \\
    precision\_of\_language & \colorvaluegradient{-0.09} & How exact and unambiguous is the wording of the choice overall? \\
    ingredient\_precision & \colorvaluegradient{-0.08} & How precise are the ingredient quantities, measurements, and specifications in the choice? \\
    beginner\_friendliness & \colorvaluegradient{-0.08} & How accessible is the choice for a novice cook? \\
    expert\_assumption & \colorvaluegradient{0.08} & How much does the choice assume the reader already understands recipe conventions and cooking basics? \\
    conversational\_tone & \colorvaluegradient{0.07} & How conversational or personable is the tone of the choice? \\
    temperature\_specificity & \colorvaluegradient{-0.06} & How specifically does the choice state cooking temperatures or heat levels? \\
    doneness\_guidance & \colorvaluegradient{0.05} & How much does the choice help the reader judge when the food is properly cooked? \\
    ingredient\_order\_clarity & \colorvaluegradient{-0.05} & How clearly does the choice present ingredients in the order they are used? \\
    yield\_information & \colorvaluegradient{0.04} & How much does the choice specify yield, servings, or portion count? \\
    alternative\_method\_coverage & \colorvaluegradient{-0.04} & How much does the choice mention alternate cooking methods or fallback approaches? \\
    optional\_variation\_content & \colorvaluegradient{-0.03} & How much does the choice include optional ingredients, substitutions, or alternative methods? \\
    ingredient\_preparation\_detail & \colorvaluegradient{-0.03} & How much detail does the choice provide about prep states such as chopped, peeled, drained, softened, or toasted? \\
    make\_ahead\_orientation & \colorvaluegradient{0.03} & How much does the choice support advance preparation, chilling, marinating, or storage planning? \\
    brand\_specificity & \colorvaluegradient{-0.02} & How strongly does the choice rely on named brands or proprietary products? \\
    action\_verb\_density & \colorvaluegradient{-0.02} & How action-oriented is the choice in its phrasing? \\
    safety\_caution\_level & \colorvaluegradient{-0.02} & How much does the choice include food safety or cautionary advice? \\
    ingredient\_contextualization & \colorvaluegradient{0.02} & How much does the choice contextualize ingredients with examples, alternatives, or explanatory labels? \\
    sensory\_cue\_usage & \colorvaluegradient{-0.01} & How much does the choice use sensory cues such as color, texture, aroma, or sound? \\
    process\_completeness & \colorvaluegradient{-0.01} & How complete is the recipe process from preparation through finishing? \\
    storage\_guidance & \colorvaluegradient{0.01} & How much does the choice explain how long or how to store the finished item or intermediate preparation? \\
    finishing\_step\_emphasis & \colorvaluegradient{0.01} & How much attention does the choice give to final steps such as garnishing, cooling, resting, glazing, or topping? \\
    specialized\_terminology & \colorvaluegradient{0.01} & How much does the choice use specialized culinary vocabulary or technical terms? \\
    instructional\_detail & \colorvaluegradient{0.00} & How much step-by-step procedural detail does the choice provide? \\
    time\_specificity & \colorvaluegradient{0.00} & How specifically does the choice state cooking, resting, marinating, or chilling times? \\
    structural\_cohesion & \colorvaluegradient{0.00} & How well organized and internally coherent is the choice as a complete recipe instruction? \\
    \bottomrule
    \end{tabular}
    }
    \caption{Features discovered by FaST on Recipe and their learned weights. Features are ordered by the magnitude of their corresponding weight in absolute value.}
    \label{tab:recipe-features40}
\end{table*}

\section{Prompts}
\label{app:prompts}

We provide here the main prompts used in our experiments. These prompts have been adapted to the CORA and Recipe datasets from the prompts originally defined in~\citet{fast}.
\begin{itemize}
    \item \textbf{FaRM-related prompts:} Table~\ref{tab:prompt-scoring} describes the prompts used for response scoring with prompted LLM-based feature functions.
    
    \item \textbf{Generation prompts:} Tables~\ref{tab:prompt-system},~\ref{tab:prompt-gen-zeroshot}, and~\ref{tab:prompt-gen-prompting} contain the system prompts, the Zeroshot prompts (also used to sample candidate responses in GRPO and RFT), and the Prompting approach prompts, respectively.

    \item \textbf{Evaluation prompts:} Table~\ref{tab:prompt-helpfulness} provides the prompts used to estimate the helpfulness of the responses via an LLM judge.
\end{itemize}


\begin{table*}[t]
\centering
\scalebox{0.96}{
\begin{tabular}{c@{\hskip 0.04\textwidth}c}
\toprule
\textbf{CORA} & \textbf{Recipe} \\
\midrule
\multicolumn{2}{c}{\textit{System prompt}}  \\
\arrayrulecolor{lightgray}\midrule
\begin{minipage}[t]{0.48\textwidth}
\small
You are a scoring assistant that evaluates responses generated by an AI coffee ordering assistant.
\end{minipage}
&
\begin{minipage}[t]{0.48\textwidth}
\small
You are a scoring assistant that evaluates responses generated by an AI cooking assistant.
\end{minipage}
\\
\midrule
\multicolumn{2}{c}{\textit{User prompt}} \\
\midrule
\begin{minipage}[t]{0.48\textwidth}
\small
You will be given a question that can be submitted to an AI coffee ordering assistant, and a response that attempts to answer this question. Your job is to rate the response based on the following criterion: \textcolor{blue}{\{attribute\_desc\}}. Score the response on a scale from 1 to 5 where 1 means \textcolor{blue}{\{attr\_min\}} and 5 means \textcolor{blue}{\{attr\_max\}}. Here are the question and the related response:

\vspace{1em}
\# Question: \textcolor{blue}{\{context\}} \\
\# Response: \textcolor{blue}{\{response\}}

\vspace{1em}
Answer by outputting a number from 1 to 5 (and nothing else).

\vspace{1em}
Score:
\end{minipage}
&
\begin{minipage}[t]{0.48\textwidth}
\small
You will be given a question that can be submitted to an AI cooking assistant, and a response that attempts to answer this question. Your job is to rate the response based on the following criterion: \textcolor{blue}{\{attribute\_desc\}}. Score the response on a scale from 1 to 5 where 1 means \textcolor{blue}{\{attr\_min\}} and 5 means \textcolor{blue}{\{attr\_max\}}. Here are the question and the related response:

\vspace{1em}
\# Question: \textcolor{blue}{\{context\}} \\
\# Response: \textcolor{blue}{\{response\}}

\vspace{1em}
Answer by outputting a number from 1 to 5 (and nothing else).

\vspace{1em}
Score:
\end{minipage}
\\
\arrayrulecolor{black}\bottomrule
\end{tabular}
}
\caption{Feature function prompts for CORA (left) and Recipe (right). These prompts are used to obtain feature-wise response scores. The feature is specified by the fields \textcolor{blue}{attribute\_desc} (overall description of the feature), \textcolor{blue}{attr\_min} (description of the minimum score) and \textcolor{blue}{attr\_max} (description of the maximum score) generated in the feature discovery step.}
\label{tab:prompt-scoring}
\end{table*}


\begin{table*}[t]
\centering
\scalebox{0.96}{
\begin{tabular}{c@{\hskip 0.04\textwidth}c}
\toprule
\textbf{CORA} & \textbf{Recipe} \\
\midrule
\multicolumn{2}{c}{\textit{System prompt}} \\
\arrayrulecolor{lightgray}\midrule

\begin{minipage}[t]{0.48\textwidth}
\small
You are an AI coffee ordering assistant that writes responses to answer user questions.

\vspace{1em}
Your responses should be based on the following information about the coffee ordering service you are managing.

\vspace{1em}
COFFEE HOUSE MENU: \\
- Espresso - Rich shot with caramel crema. \$3.25 (double shot +\$1.00) \\
- Americano - Espresso topped with hot water. \$3.50 \\
- Latte - Smooth espresso with steamed milk; flavors: vanilla, hazelnut, caramel. \$4.75 \\
- Cappuccino - Equal parts espresso, steamed milk, and foam dusted with cocoa. \$4.50 \\
- Flat White - Velvety espresso with micro-foamed milk. \$4.25 \\
- Cold Brew - 18-hour steeped coffee served over ice; add oat milk or sweet cream. \$4.95 \\
- Mocha - Espresso, chocolate syrup, steamed milk, whipped cream finish. \$5.00 \\
- Tea Latte - Breakfast black tea with steamed oat milk and honey. \$4.15 \\
- Bakery Pairings - Fresh bakes delivered at 6am daily: \\
\hspace*{1em}- Almond croissant with toasted almonds. \$3.95 \\
\hspace*{1em}- Blueberry muffin with lemon zest glaze. \$3.25 \\
\hspace*{1em}- Banana bread slice with brown butter drizzle. \$3.45 \\
- Seasonal Special - Cardamom maple latte with cinnamon dust. \$5.25

\vspace{1em}
Store hours: 6:00 AM to 8:00 PM daily \\
Milk alternatives: oat milk, almond milk, soy milk (add \$0.50) \\
Loyalty program: Sign up at coffeehouse.com/rewards \\
Order ahead: Use app or visit coffeehouse.com/order \\
Contact: support@coffeehouse.com or call 555-CAFE
\end{minipage}
&
\begin{minipage}[t]{0.48\textwidth}
\small
You are an AI cooking assistant that writes recipe descriptions to answer user cooking queries.
\end{minipage}
\\

\arrayrulecolor{black}\bottomrule
\end{tabular}
}
\caption{System prompts used for generation on CORA (left) and Recipe (right).}
\label{tab:prompt-system}
\end{table*}


\begin{table*}[t]
\centering
\scalebox{0.96}{
\begin{tabular}{c@{\hskip 0.04\textwidth}c}
\toprule
\textbf{CORA} & \textbf{Recipe} \\
\midrule
\multicolumn{2}{c}{\textit{Zeroshot user prompt}} \\
\arrayrulecolor{lightgray}\midrule

\begin{minipage}[t]{0.48\textwidth}
\small
You will be given a question asked by a user of a coffee ordering service. Your job is to write a response using the style of your choice (including wording and formatting). The length of your response can range from a single sentence to a short paragraph. Do not include any introduction, preamble, explanation or conclusion - only the direct response to the question.

\vspace{1em}
Here is the question:

\vspace{1em}
\# Question: \textcolor{blue}{\{context\}} \\
\# Response:
\end{minipage}
&
\begin{minipage}[t]{0.48\textwidth}
\small
You will be given a query from a user asking how to prepare a certain recipe. Your job is to write the steps of the requested recipe preparation using the style of your choice (including wording and formatting). The length of your response can range from a single sentence to a short paragraph. Do not include any introduction, preamble, explanation, conclusion or list of ingredients/utensils - only the description of the recipe steps.

\vspace{1em}
Here is the user query:

\vspace{1em}
\# Question: \textcolor{blue}{\{context\}} \\
\# Response:
\end{minipage}
\\

\arrayrulecolor{black}\bottomrule
\end{tabular}
}
\caption{Zeroshot generation prompts for CORA (left) and Recipe (right). These prompts are used to generate responses in the Zeroshot approach and sample candidate responses for GRPO and RFT.}
\label{tab:prompt-gen-zeroshot}
\end{table*}


\begin{table*}[t]
\centering
\scalebox{0.96}{
\begin{tabular}{c@{\hskip 0.04\textwidth}c}
\toprule
\textbf{CORA} & \textbf{Recipe} \\
\midrule
\multicolumn{2}{c}{\textit{Prompting approach user prompt}} \\
\arrayrulecolor{lightgray}\midrule

\begin{minipage}[t]{0.48\textwidth}
\small
You will be given some requirements on the response formulation, and a question. Your job is to write a response to the question while aligning as closely as possible with the provided requirements. The length of your response can range from a single sentence to a short paragraph. Do not include any introduction, preamble, explanation or conclusion - only the direct response to the question.

\vspace{1em}
Here are the requirements on the response formulation:

\vspace{1em}
\textcolor{blue}{\{profile\_desc\}}

\vspace{1em}
Here is the question:

\vspace{1em}
\# Question: \textcolor{blue}{\{context\}} \\
\# Response:
\end{minipage}
&
\begin{minipage}[t]{0.48\textwidth}
\small
You will be given some requirements on the response formulation, and a user cooking query asking how to prepare a certain recipe. Your job is to write the steps of the requested recipe preparation while aligning as closely as possible with the provided requirements. The length of your response can range from a single sentence to a short paragraph. Do not include any introduction, preamble, explanation, conclusion or list of ingredients/utensils - only the description of the recipe steps.

\vspace{1em}
Here are the requirements on the response formulation:

\vspace{1em}
\textcolor{blue}{\{profile\_desc\}}

\vspace{1em}
Here is the user query:

\vspace{1em}
\# Question: \textcolor{blue}{\{context\}} \\
\# Response:
\end{minipage}
\\

\arrayrulecolor{black}\bottomrule
\end{tabular}
}
\caption{Prompting approach generation prompts for CORA (left) and Recipe (right). These prompts condition response generation on TTS-friendliness requirements listed in \textcolor{blue}{profile\_desc}, detailed in Table~\ref{tab:profile-desc}.}
\label{tab:prompt-gen-prompting}
\end{table*}

\begin{table*}[t]
\centering
\scalebox{0.92}{
\begin{tabular}{p{0.95\textwidth}}
\toprule
\textbf{TTS-friendliness requirements for the Prompting approach} \\
\midrule

The text response should be suitable and understandable if this response is uttered to a human user via Text-to-Speech. In other words, the response should be ``speech-friendly''. Here is a list of rules describing what constitutes a speech-friendly text: \\

R1: Prefer short, simple sentences \\
R2: One main idea per sentence \\
R3: Use explicit discourse markers (``first'', ``however'') \\
R4: Avoid references to written form (``see above'', ``as shown below'') \\
R5: Prefer common, easily pronounced words \\
R6: Avoid long sequences of similar sounds / tongue twisters \\
R7: Limit code-switching / foreign terms \\
R8: Expand Latin abbreviations (e.g., i.e., etc.) \\
R9: Write numbers how they should be spoken \\
R10: Make dates unambiguous (``March 4th, 2025'') \\
R11: Spell out units and symbols instead of \%, \textdegree C, €, \&, \# \\
R12: Avoid dense numeric strings / long IDs \\
R13: Use standard punctuation only (no ``?!?!'' etc.) \\
R14: Use commas for natural pauses, not decoration \\
R15: Use ? and ! correctly and sparingly \\
R16: Avoid raw URLs and email addresses \\
R17: Avoid code / markup in normal utterances \\
R18: Avoid emojis and emoticons \\
R19: Avoid visually structured lists; use spoken lists instead \\
R20: Use conversational, speech-like style \\
R21: Avoid abrupt context switches mid-utterance \\
R22: Avoid nested quotations where possible \\
R23: Minimize ambiguous homographs in same sentence (``lead/lead'') \\
R24: Avoid creative spacing/caps/repetition (``weIrD'', ``soooo'', ``l o n g'') \\
R25: Keep language consistent within utterance \\

\bottomrule
\end{tabular}
}
\caption{TTS-friendliness requirements for the Prompting approach, used as the \textcolor{blue}{profile\_desc} field in the user prompt (see Table~\ref{tab:prompt-gen-prompting}).}
\label{tab:profile-desc}
\end{table*}

\begin{table*}[t]
\centering
\scalebox{0.95}{
\begin{tabular}{c@{\hskip 0.03\textwidth}c}
\toprule
\textbf{Reference-free judge (CORA)} & \textbf{Reference-based judge (Recipe)} \\
\midrule

\begin{minipage}[t]{0.47\textwidth}
\small
You are evaluating whether an assistant's response adequately addresses a user's question.

\vspace{0.5em}
You will be given: \\
- A user question (CONTEXT) \\
- A generated answer (GENERATED) to evaluate

\vspace{0.5em}
Follow these steps before scoring:

\vspace{0.5em}
STEP 1 - Identify the key requests or questions the user is making in CONTEXT (there may be one or several). \\
STEP 2 - For each request, determine whether GENERATED plausibly addresses it. You are NOT checking factual accuracy - only whether the answer attempts to respond to each request. \\
STEP 3 - Assign a score 1-5 based on overall coverage and coherence.

\vspace{0.5em}
5 - Excellent: All user requests are addressed, answer is coherent and complete \\
4 - Good: Most requests addressed, minor omission or slight tangent \\
3 - Acceptable: Some requests addressed but at least one clearly missed or only vaguely touched \\
2 - Poor: Most requests ignored or the answer is largely off-topic \\
1 - Bad: Complete non-sequitur, gibberish, self-contradictory, or obviously broken output

\vspace{0.5em}
CRITICAL RULES: \\
- A response that sounds natural and fluent but ignores part of the question must score LOW \\
- A response that sounds robotic or unnatural but addresses all requests must score HIGH \\
- Do NOT reward or penalize based on style, tone, formatting, or how ``speech-friendly'' the answer sounds \\
- Do NOT penalize for factual errors - you cannot verify facts without a reference \\
- Only ask: does the generated answer attempt to address ALL of the user's requests?

\vspace{0.5em}
CONTEXT: \textcolor{blue}{\{context\}} \\
GENERATED: \textcolor{blue}{\{generated\_choice\}}

\vspace{0.5em}
STEP 1 - User requests: $<$list the key requests, one per line$>$ \\
STEP 2 - Coverage: $<$for each request: addressed / partially addressed / not addressed$>$ \\
SCORE: $<$single integer 1-5$>$
\end{minipage}
&
\begin{minipage}[t]{0.47\textwidth}
\small
You are evaluating whether an assistant's response correctly and helpfully answers a user question.

\vspace{0.5em}
You will be given: \\
- A user question (CONTEXT) \\
- A reference answer (REFERENCE) considered correct and helpful \\
- A generated answer (GENERATED) to evaluate

\vspace{0.5em}
Rate the GENERATED answer on a scale from 1 to 5 based SOLELY on factual correctness and helpfulness:

\vspace{0.5em}
5 - Excellent: Fully and correctly answers the question, consistent with the reference \\
4 - Good: Mostly correct, minor omissions or slight differences, nothing misleading \\
3 - Acceptable: Partially answers the question, some inaccuracies or irrelevant additions \\
2 - Poor: Mostly incorrect or misses the point, significant factual divergence from reference \\
1 - Bad: Wrong, completely off-topic, or introduces false/harmful information

\vspace{0.5em}
CRITICAL RULES: \\
- A response that sounds natural and fluent but gives WRONG or INCOMPLETE information must score LOW \\
- A response that sounds robotic or unnatural but gives CORRECT and COMPLETE information must score HIGH \\
- Do NOT reward or penalize based on style, tone, formatting, or how ``speech-friendly'' the answer sounds \\
- Only ask: does the generated answer correctly and helpfully respond to the user's question?

\vspace{0.5em}
CONTEXT: \textcolor{blue}{\{context\}} \\
REFERENCE: \textcolor{blue}{\{preferred\_choice\}} \\
GENERATED: \textcolor{blue}{\{generated\_choice\}}

\vspace{0.5em}
REASONING: -$<$one sentence explaining your score$>$ \\
SCORE: $<$single integer 1-5$>$
\end{minipage}
\\

\bottomrule
\end{tabular}
}
\caption{Prompts used for helpfulness evaluation. The reference-free judge (left) evaluates whether the generated response addresses the user's requests without relying on factual verification, while the reference-based judge (right) evaluates correctness and helpfulness relative to a reference answer. The former is used for the CORA dataset, while the latter is used for the Recipe dataset.}
\label{tab:prompt-helpfulness}
\end{table*}

\end{document}